\documentclass[10pt,twocolumn,letterpaper]{article}

\usepackage{iccv}
\usepackage{times}
\usepackage{epsfig}
\usepackage{graphicx}
\usepackage{amsmath}
\usepackage{amssymb}

\usepackage{xspace}

\usepackage{multirow}
\usepackage{colortbl}
\usepackage{tabularx}
\usepackage{booktabs}
\usepackage[table]{xcolor}
\usepackage{caption}
\usepackage{tablefootnote}
\usepackage{url}

\usepackage{enumitem}
\setitemize{noitemsep,topsep=0pt,parsep=0pt,partopsep=0pt, leftmargin=12pt}

\usepackage{xspace}
\usepackage{bbold}
\usepackage{xcolor}

\newcommand{\vct}[1]{\ensuremath{\boldsymbol{#1}}}
\newcommand{\mat}[1]{\mathtt{#1}}
\newcommand{\set}[1]{\ensuremath{\mathcal{#1}}}
\newcommand{\con}[1]{\ensuremath{\mathsf{#1}}}
\newcommand{\T}{\ensuremath{^\top}}

\newcommand{\iABNsync}{iABN$^\text{sync}$\xspace}
\newcommand{\iABN}{iABN\xspace}
\newcommand{\convx}[1]{\textrm{conv#1}}
\newcommand{\monodis}{MonoDIS}
\newcommand{\ioudis}{IoUDIS}
\newcommand{\confidence}{3DConf}

\definecolor{mapillarygreen}{RGB}{5,203,99}

\usepackage[pagebackref=true,breaklinks=true,letterpaper=true,colorlinks,bookmarks=false]{hyperref}

\renewcommand{\paragraph}[1]{

        \vspace{3pt}
	\noindent\textbf{#1}}

\newcommand*{\dotcompass}{\ensuremath\vcenter{\hbox{\includegraphics[width=.5em]{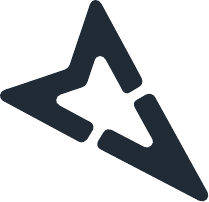}}}} 

\iccvfinalcopy 

\def\iccvPaperID{3921} 

\ificcvfinal\fi
\begin{document}

\title{
\vspace{-20pt}
Disentangling Monocular 3D Object Detection\\
}

\author{Andrea Simonelli$^{\dotcompass,\star}$, Samuel Rota Bul\`o$^{\dotcompass}$, Lorenzo Porzi$^{\dotcompass}$, Manuel L\'opez-Antequera$^{\dotcompass}$, Peter Kontschieder$^{\dotcompass}$\\
$^{\dotcompass}$Mapillary Research, $^\star$University of Trento \\
{\tt\small research@mapillary.com}\\
}

\twocolumn[{%
\renewcommand\twocolumn[1][]{#1}%
\maketitle

\vspace{-30pt}
\begin{center}
    \includegraphics[width=.9\textwidth]{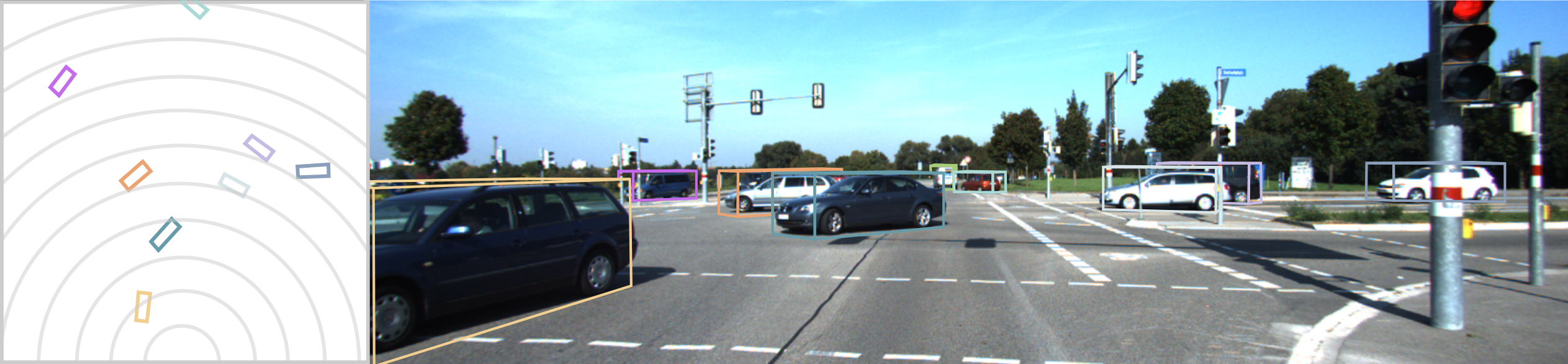}
    \captionof{figure}{Results obtained from our single image, monocular 3D object detection network \textit{MonoDIS} on a KITTI3D test image with corresponding birds-eye view, showing its ability to estimate size and orientation of objects at different scales.}
    \label{fig:catchy}
\end{center}%
}]

\maketitle

\begin{abstract}
In this paper we propose an approach for monocular 3D object detection from a single RGB image, which leverages a novel disentangling transformation for 2D and 3D detection losses and a novel, self-supervised confidence score for 3D bounding boxes. Our proposed loss disentanglement has the twofold advantage of simplifying the training dynamics in the presence of losses with complex interactions of parameters, and sidestepping the issue of balancing independent regression terms. Our solution overcomes these issues by isolating the contribution made by groups of parameters to a given loss, without changing its nature. We further apply loss disentanglement to another novel, signed Intersection-over-Union criterion-driven loss for improving 2D detection results. Besides our methodological innovations, we critically review the AP metric used in KITTI3D, which emerged as the most important dataset for comparing 3D detection results. We identify and resolve a flaw in the 11-point interpolated AP metric, affecting all previously published detection results and particularly biases the results of monocular 3D detection. We provide extensive experimental evaluations and ablation studies on the KITTI3D and nuScenes datasets, setting new state-of-the-art results on object category car by large margins.
\end{abstract}

\section{Introduction}
Recent developments in object recognition~\cite{Liu_2018_DetSurvey} have led to near-human performance on monocular 2D detection tasks. For applications with given, realistic accuracy requirements or constraints on computational budget, it is possible to choose general-purpose 2D object detectors from a large pool~\cite{Ren+15,Liu2016,Redmon2016,Wu_2017_CVPR_Workshops,Redmon_2017_CVPR,Lin+17,Law_2018_ECCV}. 

The performance situation considerably changes in the 3D object detection case. Even though there are promising methods based on multi-sensor fusion (usually exploiting LIDAR information~\cite{Liang_CVPR_2019,Wang_arxiv_2019,Shin_arxiv_18,shi2018pointrcnn} next to RGB images), 3D detection results produced from a single, monocular RGB input image lag considerably behind. This can be attributed to the ill-posed nature of the problem, where a lack of explicit knowledge about the unobserved depth dimension introduces ambiguities in 3D-to-2D mappings and hence significantly increases the task complexity. 

To still enable 3D object detection from monocular images, current works usually make assumptions about the scene geometry, camera setup or the application (\eg that cars cannot fly~\cite{qin2019monogrnet}). The implementation of such priors determines the encoding of extent and location/rotation of the 3D boxes, the corresponding 2D projections or their 3D box center depths. The magnitudes of these parameters have different units and therefore non-comparable meanings, which can negatively affect the optimization dynamics when error terms based on them are directly combined in a loss function. As a consequence, state-of-the-art, CNN-based monocular 3D detection methods~\cite{Manhardt_2019_CVPR,qin2019monogrnet} report to train their networks in a stage-wise way. First the 2D detectors are trained until their performance stabilizes, before 3D reasoning modules can be integrated. While stage-wise training \textit{per se} is not unusual in the context of deep learning, it could be an indication that currently used loss functions are yet sub-optimal. 

A significant amount of recent works are focusing their experimental analyses on the KITTI3D dataset~\cite{Geiger2012CVPR}, and in particular its \textit{Car} category~\cite{Manhardt_2019_CVPR,qin2019monogrnet,Roddick18,Xu_2018_CVPR}. The availability of suitable benchmark datasets confines the scope of experimental analyses and when only few datasets are available, progress in the research field is strongly tied to the expressiveness of used evaluation metrics. KITTI3D adopted the \textit{11-point Interpolated Average Precision} metric~\cite{Salton1986} used in the PASCAL VOC2007~\cite{Everingham2010} challenge. We found a major flaw in the metric where using a single, confident detection result per difficulty category (KITTI3D distinguishes between \textit{easy}, \textit{moderate} and \textit{hard} samples) suffices to obtain AP scores of $\approx9\%$ on a dataset level, which is up to 3$\times$ higher than the performance reported by recent works~\cite{NIPS2015_Chen,Chen_2016_CVPR,TongHe_2019_arxiv,Xu_2018_CVPR}. 

The contributions of our paper disentangle the task of monocular 3D object detection at several levels. Our major technical contribution \textit{disentangles} dependencies of different parameters by isolating and handling parameter groups individually at a loss-level. This overcomes the issue of non-comparability for parameter magnitudes, while preserving the nature of the final loss. Our loss disentanglement significantly improves losses on both, 2D and 3D tasks. It also enables us to effectively train the entire CNN architecture (2D+3D) together and end-to-end, without the need of hyperparameter-sensitive, stage-wise training or warm-up phases. As additional contributions we i) leverage 2D detection performance through a novel loss based on a \textit{signed Intersection-over-Union} criterion and ii) introduce a loss term for predicting detection confidence scores of 3D boxes, learned in a self-supervised way.

Another major contribution is a critical review of the 3D metrics used to judge progress in monocular 3D object detection, with particular focus on the predominantly used KITTI3D dataset. We observe that a flaw in the definition of the 11-point, interpolated AP metric significantly biases 3D detection results at the performance level of current state-of-the-art methods. Our applied correction, despite bringing \textit{all works evaluating on KITTI3D} back down to earth, more adequately describes their true performance. 

For all our contributions, we provide ablation studies on the KITTI3D and the novel nuScenes~\cite{Cae+19} driving datasets. Fair comparisons indicate that our work considerably improves over current monocular 3D detection methods.

\section{Related Work}
We review the most recent, related works from 3D object detection and group them according to the data modalities used therein. After discussing RGB-only works just like ours, we list works exploiting also depth and/or synthetic data augmentation or 3D shape information, before finalizing with a high-level summary about LIDAR and/or stereo-based approaches.

\paragraph{RGB images only.} 
Deep3DBox~\cite{Mousavian_2017_CVPR} proposed to estimate full 3D pose and object dimensions from a 2D box by exploiting constraints from projective geometry. The core idea is that the perspective projection of a 3D bounding box should fit tightly to at least one side of its corresponding 2D box detection. In SSD-6D~\cite{Kehl_2017_ICCV} an initial 2D detection hypothesis is lifted to provide 6D pose of 3D objects by using structured discretizations of the full rotational space. 3D model information is learned by only training from synthetically augmented datasets. OFTNet~\cite{Roddick18} introduces an orthographic feature transform, mapping features extracted from 2D to a 3D voxel map. The voxel map's features are eventually reduced to 2D (birds-eye view) by integration along the vertical dimension, and detection hypotheses are efficiently processed by exploiting integral-image representations. Mono3D~\cite{Chen_2016_CVPR} emphasized on generation of 3D candidate boxes, scored by different features like class semantics, contour, shape and location priors. Even though at test time the results are produced based on single RGB images only, their method also requires semantic and instance segmentation results as input. The basic variant (w/o using depth) of ROI-10D~\cite{Manhardt_2019_CVPR} proposes a novel loss to lift 2D detection, orientation and scale into 3D space that can be trained in an end-to-end fashion. FQNet~\cite{Liu+19} infers a fitting quality criterion in terms of 3D IoU scores, allowing them to filter estimated 3D box proposals based on using only 2D object cues. MonoGRNet~\cite{qin2019monogrnet} is the current state-of-the-art for RGB-only input, using a CNN comprised of four sub-networks for 2D detection, instance depth estimation, 3D location estimation and local corner regression, respectively. The three latter sub-networks emphasize on geometric reasoning, \ie instance depth estimation predicts the central 3D depth of the nearest object instance, 3D location estimation seeks for the 3D bounding box center by exploiting 3D to 2D projections at given instance depth estimations, and local corner regression directly predicts the eight 3D bounding box corners in a local (or allocentric~\cite{Kundu_2018_CVPR,Manhardt_2019_CVPR} way). It is relevant to mention that~\cite{qin2019monogrnet} reports that training was conducted stage-wise: First, the backbone is trained together with the 2D detector using Adam. Next, the geometric reasoning modules are trained (also with Adam). Finally, the whole network is trained end-to-end using stochastic gradient descent. The work in~\cite{Barabanau_arXiv_2019} learns to estimate correspondences between detected 2D keypoints and 3D counterparts. However, this requires manual annotations on the surface of 3D CAD models and is limited in dealing with occluded objects.

\paragraph{Including depth.} 
An expansion stage of ROI-10D~\cite{Manhardt_2019_CVPR} takes advantage of depth information provided by SuperDepth~\cite{Pillai_2019_ICRA}, which itself is learned in a self-supervised manner. In~\cite{Xu_2018_CVPR}, a multi-level fusion approach is proposed, exploiting disparity estimation results from a pre-trained module during both, the 2D box proposal generation stage as well as the 3D prediction part of their network. 

\paragraph{Including 3D shape information.}
3D-RCNN~\cite{Kundu_2018_CVPR} exploits the idea of using inverse graphics for instance-level, amodal 3D shape and pose estimation of all object instances per image. They propose a differentiable Render-and-Compare loss, exploiting available 2D annotations in existing datasets for guiding optimization of 3D object shape and pose. In~\cite{Zia_2014_CVPR}, the recognition task is tackled by jointly reasoning about the 3D shape of multiple objects. Deep-MANTA~\cite{Chabot_2017_CVPR} uses 3D CAD models and annotated 3D parts in a coarse-to-fine localization process. The work in~\cite{Murthy_17_ICRA} encodes shape priors using keypoints for recovering the 3D pose and shape of a query object. In Mono3D++~\cite{TongHe_2019_arxiv}, the 3D shape and pose for cars is provided by using a morphable wireframe, and it optimizes projection consistency between generated 3D hypotheses and corresponding, 2D pseudo-measurements.

\paragraph{LIDAR and/or stereo-based.}
3DOP~\cite{NIPS2015_Chen} exploits stereo images and prior knowledge about the scene to directly reason in 3D. Stereo R-CNN~\cite{cvpr19stereorcnn} tackles 3D object detection by exploiting stereo imagery and produces stereo boxes, keypoints, dimensions and viewpoint angles, summarized in a learned 3D box estimation module. In MV3D~\cite{Chen_2017_CVPR}, a sensor-fusion approach for LIDAR and RGB images is presented, approaching 3D object proposal generation and multi-view feature fusion via individual sub-networks. Conversely, Frustrum-PointNet~\cite{Qi_2018_CVPR} directly operates on LIDAR point clouds and aligns candidate points provided from corresponding 2D detections for estimating the final, amodal 3D bounding boxes. PointRCNN~\cite{shi2018pointrcnn} describes a 2-stage framework where the first stage provides bottom-up 3D proposals and the second stage refines them in canonical coordinates. RoarNet~\cite{Shin_arxiv_18} applies a 2D detector to first estimate 3D poses of objects from a monocular image before processing corresponding 3D point clouds to obtain the final 3D bounding boxes. 

\section{Task Description}

We address the problem of monocular 3D object detection, where the input is a single RGB image and the output consists in a 3D bounding box, expressed in camera coordinates, for each object that is present in the image (see, Fig.~\ref{fig:catchy}).
As opposed to other methods in the literature, we do \emph{not} take additional information as input like depth obtained from LIDAR or other supervised or self-supervised monocular depth estimators. Also the training data consists solely of RGB images with corresponding annotated 3D bounding boxes. Nonetheless, we require a calibrated setting so we assume that per-image calibration parameters are available both at training and test time.

\section{Proposed Architecture}

We adopt a two-stage architecture that shares a similar structure with the state-of-the-art~\cite{Manhardt_2019_CVPR}. It consists of a single-stage 2D detector (\emph{first stage}) with an additional 3D detection head (\emph{second stage}) constructed on top of features pooled from the detected 2D bounding boxes. Details of the architecture are given below.

\subsection{Backbone}
\begin{figure}[th]
    \centering
    \includegraphics[width=.8\columnwidth]{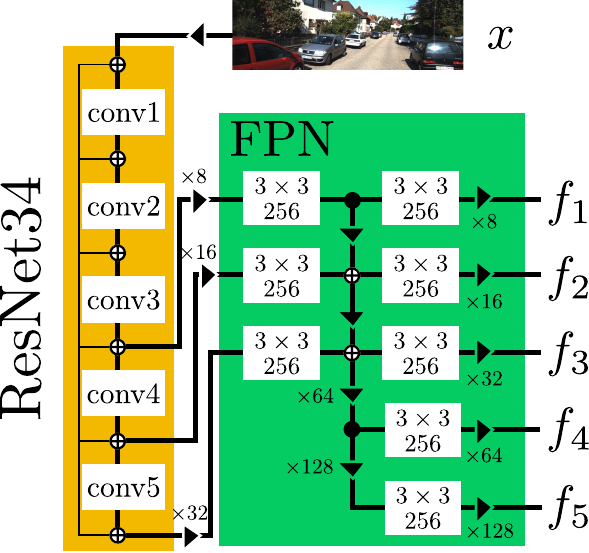}
    \caption{Backbone architecture. Rectangles in the ``FPN'' block represent convolutions followed by \iABNsync.}
    \label{fig:backbone}
    \vspace{-12pt}
\end{figure}

The backbone we use is a ResNet34~\cite{He2015b} with a Feature Pyramid Network (FPN)~\cite{Lin2016} built on top of it. The FPN network has the same structure as in~\cite{Lin+17} with 3+2 scales, connected to the output of modules \convx3, \convx4 and \convx5 of ResNet34, corresponding to downsampling factors of $\times 8$, $\times 16$ and $\times 32$, respectively. Our ResNet34 differs from the standard one by replacing BatchNorm+ReLU layers with the synchronized version of InPlaceABN (\iABNsync) activated with LeakyReLU with negative slope $0.01$ as proposed in~\cite{RotPorKon18a}. This modification does not affect the performance of the network, but allows to free up a significant amount of GPU memory, which can be exploited to scale up the batch size or input resolution. All FPN blocks depicted in Fig.~\ref{fig:backbone} correspond to $3\times 3$ convolutions with $256$ channels, followed by \iABNsync.

\paragraph{Inputs.}
The input $x$ to the backbone is a single RGB image.

\paragraph{Outputs.}
The backbone provides $5$ output tensors $\{f_1, \ldots, f_5\}$ corresponding to the $5$ different scales of the FPN network, covering downsampling factors of $\times 8$, $\times 16$, $\times 32$, $\times 64$, and $\times 128$, each with $256$ feature channels (see, Fig.~\ref{fig:backbone}).

\subsection{2D Detection Head}
\label{sec:head2D}
\begin{figure}
    \centering
    \includegraphics[width=.8\columnwidth]{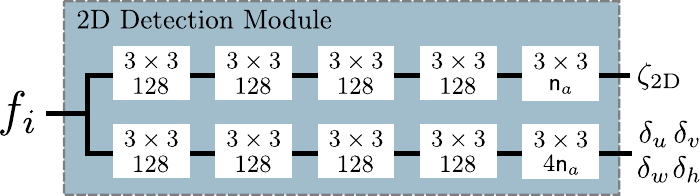}
    \caption{2D detection module. Rectangles represent convolutions. All convolutions but the last per row are followed by \iABNsync.}
    \label{fig:head2d}
    \vspace{-12pt}
\end{figure}

We consider the head of the single-stage 2D detector implemented in RetinaNet~\cite{Lin+17}, which applies a detection module independently to each output $f_i$ of the backbone described above. The detection modules share the same parameters but work inherently at different scales, according to the scale of the features that they receive as input. As opposed to the standard RetinaNet, we employ \iABNsync also in this head.
The head, depicted in Fig.~\ref{fig:head2d}, is composed of two parallel stacks of $3\times 3$ convolutions, and is parametrized by $\con n_a$ reference bounding box sizes (anchors) per scale level.

\paragraph{Inputs.}
The inputs are the $5$ outputs $\{f_1, \ldots, f_5\}$ of the backbone, where $f_i$ has a spatial resolution of $\con h_i\times \con w_i$.

\paragraph{Outputs.}
For each image, and each input tensor $f_i$, the 2D detection head generates $\con n_a$ bounding box proposals (one per anchor) for each spatial cell $g$ in the $\con h_i\times \con w_i$ grid. Each proposal for a given anchor $a$ with size $(w_a, h_a)$ is encoded as a $5$-tuple $(\zeta_{\mat {2D}}, \delta_u, \delta_v, \delta_w, \delta_h)$ such that 
\begin{itemize}
\item $p_{\mat{2D}}=(1+e^{-\zeta_\mat{2D}})^{-1}$ gives the confidence of the 2D bounding box prediction,
\item $(u_b,v_b)=(u_g+\delta_u w_a,v_g+\delta_v h_a)$ gives the center of the bounding box with $(u_g,v_g)$ being the image coordinates of cell $g$, and
\item $(w_b,h_b)=(w_a e^{\delta_w}, h_a e^{\delta_h})$ gives the bounding box size.
\end{itemize}
Fig.~\ref{fig:notation} gives a visual description of the head's outputs.

\paragraph{Losses.}
We employ the focal loss~\cite{Lin+17} to train the bounding box confidence score. This loss takes the following form, for a given cell $g$ and anchor $a$ with target confidence $y\in\{0,1\}$ and predicted confidence $p\in[0,1]$:
\[
L_\mathtt{2D}^\text{conf}(p_\mathtt{2D},y) = -\alpha y(1-p_\mathtt{2D})^\gamma \log p_\mathtt{2D} - \bar\alpha \bar y p_\mathtt{2D}^\gamma\log(1-p_\mathtt{2D})\,,
\]
where $\alpha\in[0,1]$ and $\gamma>0$ are hyperparameters that modulate the importance of errors and positives, respectively, $\bar \alpha=1-\alpha$ and $\bar y=1-y$.
The confidence target $y$ does not depend on the regressed bounding box, but only on the cell $g$ and the anchor $a$. It takes value $1$ if the reference bounding box centered in $(u_g,v_g)$ with size $(w_a,h_a)$ exhibits an Intersection-over-Union (IoU) with a ground-truth bounding box larger than a given threshold $\tau_\text{iou}$.
For each cell $g$ and anchor $a$ that matches a ground-truth bounding box $\vct{\hat b}$
with predicted bounding box $\vct b=(u_b-\frac{w_b}{2}, v_b-\frac{h_b}{2}, u_b+\frac{w_b}{2}, v_b+\frac{h_b}{2})$
we consider the following detection loss:
\begin{equation}\label{eq:det2d}
L_\mathtt{2D}^\text{bb}(\vct b, \vct {\hat b})=1-\text{sIoU}(\vct b,\vct{\hat b})\,,
\end{equation}
where $\text{sIoU}$ represents an extension of the common IoU function, which prevents gradients from vanishing in case of non-overlapping bounding boxes. 
We call it \emph{signed} IoU function, as, intuitively, it creates negative intersections in case of disjoint bounding boxes (see, Appendix~\ref{sec:siou}).
In Sec.~\ref{sec:disentangling}, we discuss a disentangling transformation of the loss in Eq.~\eqref{eq:det2d} that allows to isolate the contribution of each network's output to the loss, while preserving the fundamental nature of the loss.

\paragraph{Output Filtering.}
The dense output of the 2D head is filtered as in~\cite{Lin+17}: first, detections with scores lower than $0.05$ are discarded, then Non-Maxima Suppression (NMS) with IoU threshold $0.5$ is performed on the $5000$ top-scoring among the remaining ones, and the best $100$ are kept.

\subsection{3D Detection Head}\label{sec:head3D}
\begin{figure}
    \centering
    \includegraphics[width=\columnwidth]{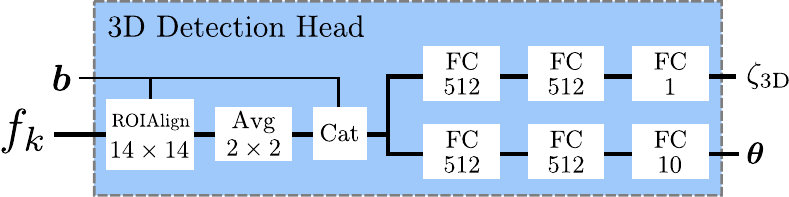}
    \caption{3D detection head. ``FC'' rectangles represent fully connected layers. All FCs except the last of each row are followed by \iABN.}
    \label{fig:head3d}
    \vspace{-12pt}
\end{figure}

The 3D detection head (Fig.~\ref{fig:head3d}) regresses a 3D bounding box for each 2D bounding box returned by the 2D detection head (surviving the filtering step). It starts by applying ROIAlign~\cite{He2017} to pool features from FPN into a $14\times 14$ grid for each 2D bounding box, followed by $2\times 2$ average pooling, resulting in feature maps with shape $7\times 7\times 128$.
The choice of which FPN output is selected for each bounding box $\vct b$ follows the same logic as in~\cite{Lin2016}, namely the features are pooled from the output $f_k$, where $k=\min(5,\max(1,\lfloor 2 + \log_2(\sqrt{w_bh_b}/224) \rfloor))$.
On top of this, two parallel branches of fully connected layers with 512 channels compute the outputs detailed below. Each fully connected layer but the last one per branch is followed by \iABN (non-synchronized).

\paragraph{Input.} The inputs are a 2D bounding box proposal $\vct b$ returned by the 2D detection head and features $f_k$ from the backbone.

\begin{figure}
    \centering
    \includegraphics[width=\columnwidth]{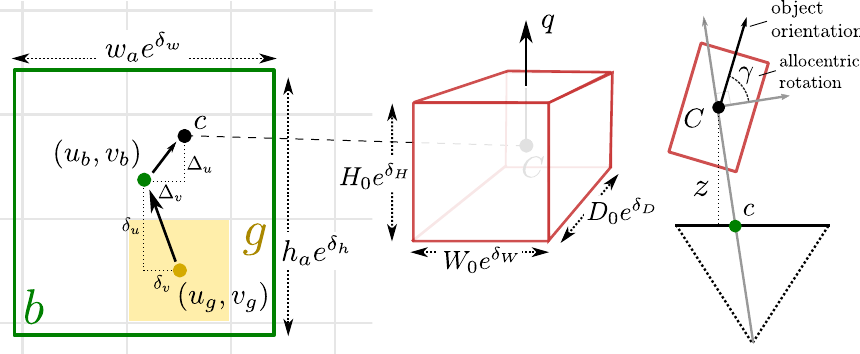}
    \caption{Visualization of the semantics of the outputs of the 2D and 3D detection heads. Left: 2D bounding box regression on image plane. Center: 3D bounding box regression. Right: allocentric angle from bird-eye view. }
    \label{fig:notation}
    \vspace{-12pt}
\end{figure}

\paragraph{Output.} The head returns for each 2D proposal $\vct b$ with center $(u_b, v_b)$ and dimensions $(w_b, h_b)$  a 3D bounding box encoded in terms of a $10$-tuple $\vct \theta=(\delta z, \Delta_u, \Delta_v, \delta_W, \delta_H, \delta_D, q_r, q_i, q_j, q_k)$ and an additional output $\zeta_\mathtt{3D}$ such that
\begin{itemize}
    \item $p_\mathtt{3D|2D}=(1+e^{-\zeta_\mat{3D}})^{-1}$ represents the confidence of the 3D bounding box prediction given the 2D proposal,
    \item $z=\mu_z+\sigma_z\delta_z$ represents the depth of the center $\vct C$ of the predicted 3D bounding box, where $\mu_z$ and $\sigma_z$ are given, dataset-wide depth statistics, 
    \item $\vct c=(u_b+\Delta_u, v_b+\Delta_v)$ gives the position of $\vct C$ projected on the image plane (in image coordinates),
    \item $\vct s=(W_0 e^{\delta_W}, H_0 e^{\delta_H},D_0 e^{\delta_D})$ is the size of the 3D bounding box, where $(W_0, H_0, D_0)$ is a given, dataset-wide reference size, and
    \item $\vct q=q_r+q_i\mat i +q_j\mat j +q_k\mat k$ is the quaternion providing the pose of the bounding box with respect to an \emph{allocentric}~\cite{Kundu_2018_CVPR}, local coordinate system.
\end{itemize}
Fig.~\ref{fig:notation} gives a visual description of the head's outputs.

\paragraph{Losses.}
Let $\theta$ be the $10$-tuple representing the regressed 3D bounding box and let $\hat B\in\mathbb R^{3\times 8}$ be the ground-truth 3D bounding box in camera coordinates.
By applying the lifting transformation $\set F$ introduced in~\cite{Manhardt_2019_CVPR} and reviewed in Appendix~\ref{sec:lifting}, we obtain the predicted 3D bounding box $B$ given the network's output $\vct \theta$, \ie $B=\set F(\vct \theta)$.
The loss on the 3D bounding box regression is then given by
\begin{equation}\label{eq:det3d}
L_\mathtt{3D}^\text{bb}(B, \hat B)=\frac{1}{8}\Vert B-\hat B\Vert_\text{H}\,,    
\end{equation}
where $\Vert\cdot\Vert_\text{H}$ denotes the Huber loss with parameter $\delta_H$ applied component-wise to each element of the argument matrix.
The loss for the confidence $p_\mathtt{3D|2D}$ about the predicted 3D bounding box is self-supervised by the 3D bounding box loss remapped into a probability range via the transformation $\hat p_\mathtt{3D|2D}=e^{-\frac{1}{T}L_\mathtt{3D}^\text{bb}(B, \hat B)}$, where $T>0$ is a temperature parameter. The confidence loss for the 3D bounding box is then the standard binary cross entropy loss:
\[
L_\text{3D}^\text{conf}(p_\mathtt{3D|2D}, \hat p_\mathtt{3D|2D})=-\hat p\log p -(1-\hat p)\log(1-p)\,,
\]
where we have omitted the subscripts for the sake of readability. This loss allows to obtain a more informed confidence about the quality of the returned 3D bounding box than just using the 2D confidence.
Akin to the 2D case, we employ also a different variant of Eq.~\eqref{eq:det3d} that disentangles the contribution of groups of parameters in order to improve the stability and effectiveness of the training. Yet, the confidence computation 
will be steered by Eq.~\eqref{eq:det3d}.

\paragraph{Output Filtering.}
The final output will be filtered based on a combination of the 2D and 3D confidences, following a Bayesian rule. The 3D confidence $p_\mathtt{3D|2D}$ is implicitly conditioned on having a valid 2D bounding box and the latter probability is reflected by $p_\mathtt{2D}$. At the same time the confidence of a 3D bounding box given an invalid 2D bounding box defaults to $0$. Hence, the unconditioned 3D confidence can be obtained by the law of total probability as
\[
p_\mathtt{3D}=p_\mathtt{3D|2D}p_\mathtt{2D}\,.
\]
This is the final confidence that our method associates to each 3D detection and that is used to filter the predictions via a threshold $\tau_\text{conf}$.
We do not perform further NMS steps on the regressed 3D bounding boxes nor filtering based on 3D prior knowledge (\eg one could reduce false positives by dropping "flying" cars).

\section{Disentangling 2D and 3D Detection Losses}\label{sec:disentangling}

In this section we propose a transformation that can be applied to the 2D bounding box loss $L_\text{2D}^{bb}$ and the 3D counterpart $L_\text{3D}^{bb}$, as well as a broader set of loss functions.
We call it \emph{disentangling} transformation because it isolates the contribution of groups of parameters to a given loss, while preserving its inherent nature. Each parameter group keeps its independent loss term, but they are all made comparable, thus sidestepping the difficulty of finding a proper weighting. While losses that combine parameters in a single term, such as those in Eq.~\eqref{eq:det2d} and Eq.~\eqref{eq:det3d}, are immune to the balancing issue, they might exhibit bad dynamics during the optimization as we will show with a toy experiment. The transformation we propose, instead, retains the best of both worlds.
\begin{figure*}[t]
    \centering
    \includegraphics[width=.48\textwidth]{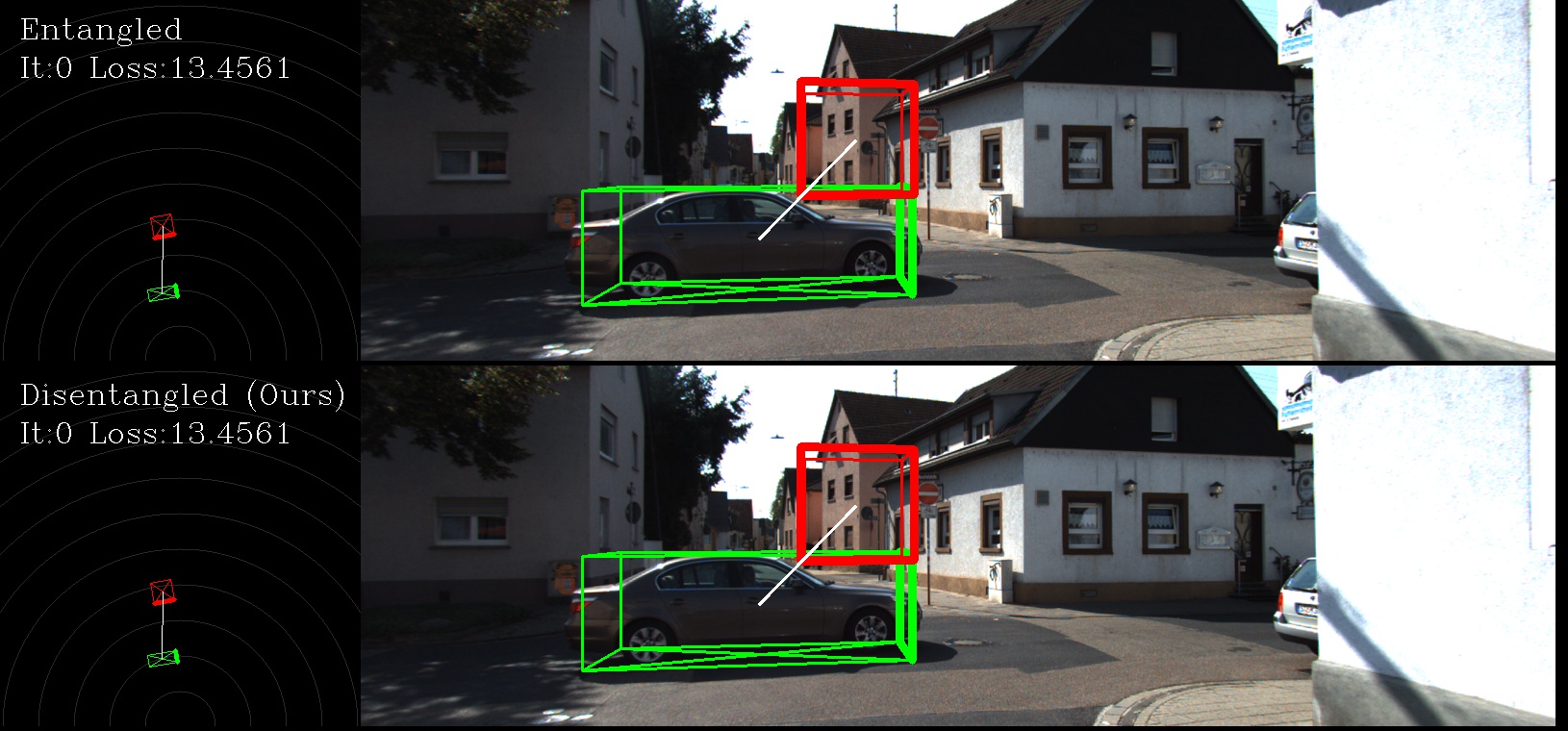}
    \includegraphics[width=.48\textwidth]{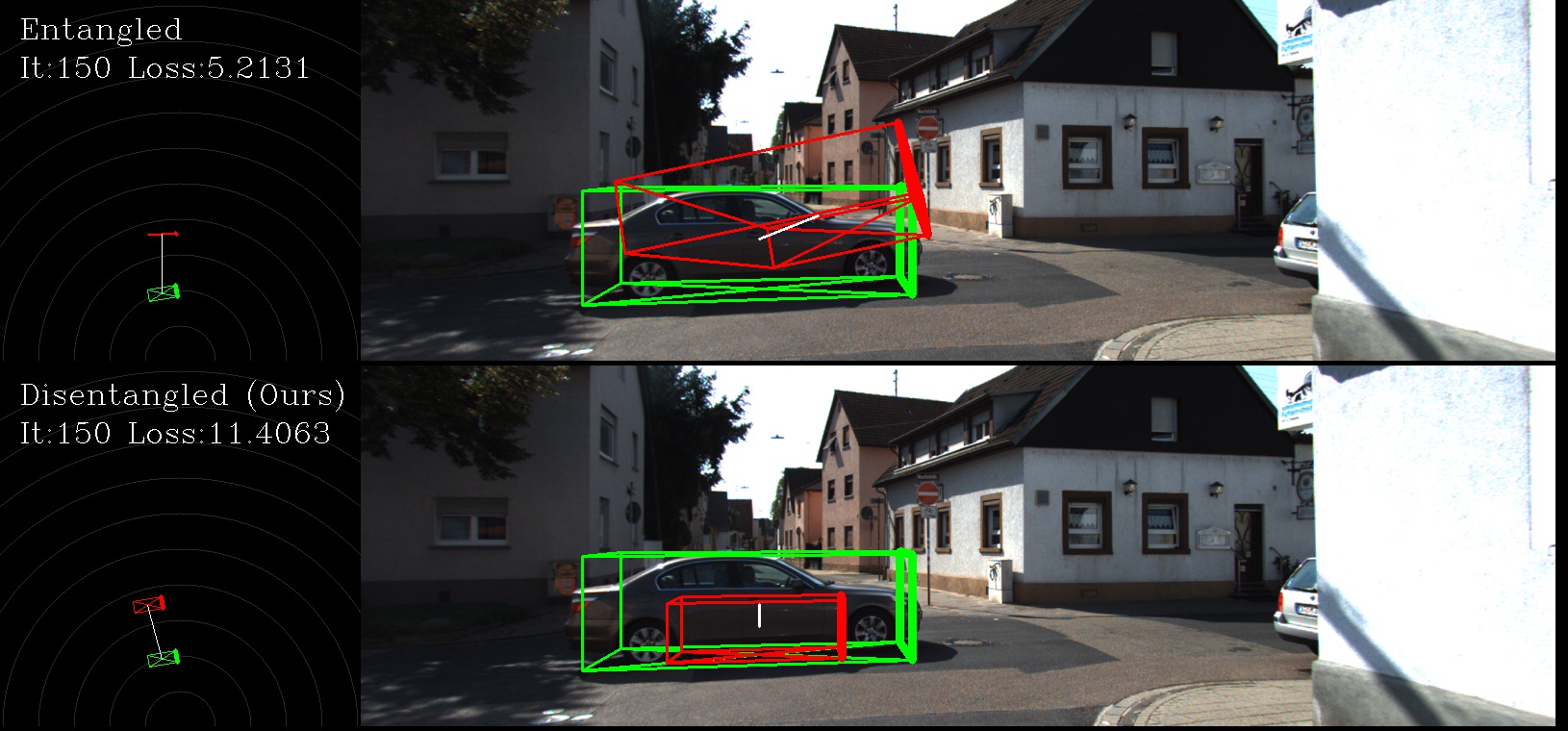}
    \caption{Sample frames from the toy experiment's video on both, entangled (top) and disentangled (bottom) runs. Optimization process at iteration $0$ (left) and $150$ (right). Green is the ground-truth target. Red is the current prediction. The face with thick lines represents the front of the car. The face with a cross represents the bottom of the car. The birds-eye view on the left shows the projection of the crossed face.}
    \label{fig:video}
    \vspace{-10pt}
\end{figure*}
\begin{figure*}
    \centering
    \includegraphics[width=\textwidth]{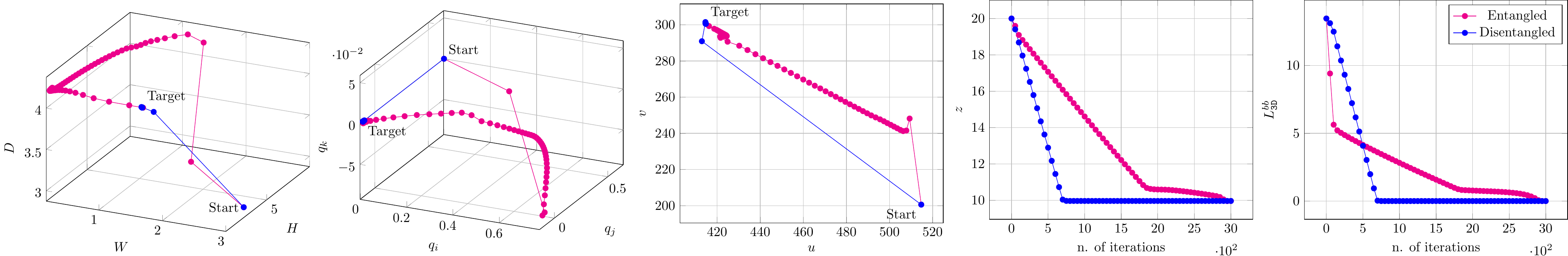}
    \caption{Trajectories of the optimization process for each group of parameters (dimensions, rotation quaternion, projected center, depth), when using the entangled (magenta) and disentangled (blue) 3D detection losses. Left-to-right: trajectories of dimensions, rotation quaternion (last 3 coordinates), projection of the 3D bounding box center on the image and depth of the 3D bounding box center. The last plot shows the evolution of the \emph{entangled} $L_\mathtt{3D}^{bb}$ loss for both cases.   }
    \label{fig:toy}
    \vspace{-10pt}
\end{figure*}

\subsection{Disentangling Transformation}
Let $L:\set Y\times\set Y\to\mathbb R_+$ be a loss function defined on a space $\set Y$ (\eg the space of 3D bounding boxes) such that $L(y,\hat y)=0$ if $\hat y=y$.
Let $\Theta\subset\mathbb R^\con d$ be a set of possible network outputs that can be mapped to elements of $\set Y$ via a function $\psi$ that we assume to be one-to-one. 
This property holds for 2D bounding boxes via the common 4D parametrization (center + dimensions), as well as for the 3D bounding boxes via the 10D representation described in Sec.~\ref{sec:head3D}.
In the latter case, $\psi$ coincides with the lifting transformation $\set F$. 
Let $\hat y$ be a fixed output element (\eg a ground-truth bounding box) and consider a partitioning of the $\con d$ dimensions of $\Theta$ into $\con k$ groups. 
To give a concrete example, in case of 2D bounding boxes we can have $2$ groups of parameters: one for the dimensions, and one for the center. In the case of $3D$ bounding boxes we consider $4$ groups related intuitively to depth, projected center, rotation and dimensions. Given $\vct\theta\in\Theta$ we denote by $\vct\theta_j$ the sub-vector corresponding to the $j$th group and by $\vct\theta_{-j}$ the sub-vector corresponding to all but the $j$th group. Moreover, given $\vct \theta,\vct \theta'\in\Theta$, we denote by $\psi(\vct \theta_j,\vct \theta'_{-j})$ the mapping of a parametrization that takes the $j$th group from $\vct \theta$ and the rest of the parameters from $\vct \theta'$.
The disentanglement of loss $L$ given $\hat y$, the mapping $\psi$ and a decomposition of parameters into $\con k$ groups is defined as:
\[
L_\text{dis}(y,\hat y)=\sum_{j=1}^\con k L(\psi(\vct \theta_j,\vct{\hat\theta}_{-j}),\hat y)\,,
\]
where $\vct\theta=\psi^{-1}(y)$ and $\vct{\hat \theta}=\psi^{-1}(\hat y)$.
The idea behind the transformation is very intuitive besides the mathematical formalism. We simply replicate $\con k$ times the loss $L$, each copy having only a group of parameters that can be optimized, the other being fixed to the ground-truth parametrization, which can be recovered via $\psi^{-1}$.
We have applied the disentangling transformation to both the 2D loss in Eq.~\eqref{eq:det2d} and to the 3D loss in Eq.~\eqref{eq:det3d} and used them to conduct our experiments, unless otherwise stated.

\subsection{Explanatory Toy Experiment}\label{sec:toy}
The toy experiment consists in comparing the optimization trajectories when we employ the (entangled) 3D object detection loss $L_\mathtt{3D}^{bb}$ and the disentangled counterpart, which is obtained by applying the disentangling transformation described in Sec.~\ref{sec:disentangling}. We took a ground-truth detection case from KITTI3D and picked an illustrative initialization for the 3D box for optimization (see, Fig.~\ref{fig:video} green and red boxes, respectively).

We perform the experiment using stochastic gradient descent with learning rate $0.001$, momentum $0.9$ and no weight decay. We run the experiment for $3000$ iterations. We report in Fig.~\ref{fig:toy} (first 4 plots from the left) the trajectories of the optimization process for each group of parameters when the entangled and disentangled losses are used. The parameter groups describe box dimensions, rotation quaternion, projected center of the 3D bounding box on the image, and the depth of the 3D bounding box center. The benefits deriving from the use of the disentangled loss can be clearly seen in the plots. Convergence is much faster and smoother. We can see that the trajectories induced by the entangled loss are suboptimal, since they explore multiple configurations of parameters before approaching the correct one, sometimes with considerable deviations (see, \eg the dimensions of the bounding box).
As an example, we report in Fig.~\ref{fig:video} (right) the point where the entangled version attains the largest deviation in terms of bounding box dimensions from the ground-truth, which happens at iteration $150$, while at this stage the optimization dynamics using the disentangled loss fixed already all parameters but the depth. Despite the quaternion being aligned with the ground-truth rotation axis from the beginning, the optimization dynamics with the entangled loss starts diverging from it, producing unnatural poses and sizes that are not properly penalized by the entangled loss as can be seen by the loss values reported for the two configurations. Such unstable supervision delivered by the entangled loss harms the generalization capabilities of the network. Interestingly, even though the optimization process that uses the disentangled loss does not directly optimize $L_\mathtt{3D}^{bb}$, it can minimize it more quickly than the counterpart directly optimizing it (see, Fig.~\ref{fig:toy} last).

We provide also a video on our project website that shows the evolution of the optimization process described above. Fig.~\ref{fig:video} gives an overview of the first frame (left column). For each optimized loss (entangled on top and disentangled on the bottom) we provide the ground truth 3D bounding box in green and the currently predicted one in red. The faces with thick lines and showing a cross represent the front of the car and the bottom of the car, respectively, while the white line connects the respective centers. We also show the birds-eye view, where we projected the bottom face (the one with the cross) on the ground plane. There we also report the value of the entangled loss $L_\mathtt{3D}^{bb}$ for both approaches for direct comparison and the iteration number. The video has been rendered with a logarithmic time scale in order to emphasize the initial part of the dynamics, which is also the most informative one.

\section{Critical Review on the KITTI3D AP Metric}
\label{sec:metric}

The KITTI3D benchmark dataset~\cite{Geiger2012CVPR} significantly determines developments and general progress on 3D object detection, and has emerged as the most decisive benchmark for monocular 3D detection algorithms like ours. It contains a total of 7481 training and 7518 test images and has no official validation set. However, it is common practice to split the training data into 3712 training and 3769 validation images as proposed in~\cite{NIPS2015_Chen}, and then report validation results. On the official test split, there is no common agreement which of the training sets to use, but in case validation data is used for snapshot cherry-picking, it is imperative to provide test data scores from the same model. 

Each 3D ground truth detection box is assigned to one out of three difficulty classes (\textit{easy, moderate, hard}), and the used 11-point Interpolated Average Precision metric is separately computed on each difficulty class. This metric was originally proposed in~\cite{Salton1986}, and was used in the PASCAL VOC challenges~\cite{Everingham2010} between 2007 and 2010. It approximates the shape of the Precision/Recall curve as
\[
	\text{AP}|_R=\frac{1}{|R|}\sum_{r\in R} \rho_{interp}(r) \,,
\]
averaging the precision values provided by $\rho_{interp}(r)$. In the current setting, KITTI3D applies exactly eleven equally spaced recall levels, \ie $R_{11}=\{0, 0.1, 0.2, \ldots, 1\}$. The interpolation function is defined as $\rho_{interp}(r)=\max\limits_{r':r'\geq r} \rho(r')$, where $\rho(r)$ gives the precision at recall $r$, meaning that instead of averaging over the actually observed precision values per point $r$, the maximum precision at recall value greater or equal than $r$ is taken. The recall intervals start at $0$, which means that a single, correctly matched prediction (according to the applied IoU level) is sufficient to obtain 100\% precision at the bottom-most recall bin. In other words, if for each difficulty level a single, but correct prediction is provided to the evaluation, this produces an $\text{AP}|_{R_{11}}$ score of $1/11\approx0.0909$ for the entire dataset, which as shown in our experimental section already outperforms a number of recent methods while it clearly does not properly assess the quality of an algorithm.

In light of KITTI3Ds importance, we propose a simple but effective fix that essentially exploits more of the information provided by the official evaluation server and evaluation scripts. Instead of sub-sampling 11 points from the provided 41 points, we approximate the area under the curve by simply replacing $R_{11}$ with $R_{40}=\{1/40, 2/40, 3/40, \ldots, 1\}$ thus averaging precision results on 40 recall positions but not at $0$. This eliminates the glitch encountered at the lowest recall bin, and allows to post-process all currently provided test server results on 2D and 3D AP scores.
\section{Experiments on KITTI3D}\label{sec:expKITTI}

We focus the validation of our method on the KITTI3D benchmark dataset that we described in Sec.~\ref{sec:metric}, using the 0.7 IoU threshold for calculating AP.

\subsection{Pre-processing}\label{ss:pre-processing}

We provide some observations about the annotations that can be found in the dataset, and some simple filtering steps that we have applied to the annotations of the training split defined in~\cite{NIPS2015_Chen}.

\paragraph{DontCare areas.} Besides standard classes such as \textit{Car}, \textit{Pedestrian} and \textit{Cyclist}, KITTI3D provides  \textit{DontCare} annotations. This class is used to label portions of the image that potentially include positive instances which have not been labeled under the proper class for reasons such as high distance. Accordingly, we avoid harvesting negatives in the 2D detection head if an anchor has IoU above $50\%$ with those areas. 

\paragraph{DontCare overlap.} Some positive bounding boxes, such as cars that were too near to the camera, have an IoU with a \textit{DontCare} bounding box greater than $50\%$. We decided to set those bounding boxes as \textit{DontCare}. This adjustment converted $729$ cars ($5.0\%$) to \textit{DontCare}.

\paragraph{Full occlusion.} Some valid bounding boxes are actually fully occluded by a nearer object. Keeping those bounding boxes as positive instances might harm the learning process, so we decided to delete them. This adjustment deleted $218$ ($1.5\%$) cars.

\vspace{3pt}
From a total number of $14357$ cars that were annotated, the valid number of \textit{Car} bounding boxes was $13410$ ($93.4\%$).

\subsection{Implementation Details}
\label{sec:details}

We give more details about our implementation and instantiation of hyperparameters, in order to enable the reproducibility of our results.

\paragraph{2D Detection Head.} For each FPN level $f_i$ and each spatial cell $g$ we employ a total of 15 anchors spanning on five aspect ratios $\{\frac{1}{3},\frac{1}{2},1,2,3\}$ and three scales $\{4 s_i 2^\frac{j}{3}\,:\,j\in\{0,1,2\}\}$, where $s_i$ is the downsampling factor of $f_i$. Each anchor is considered positive if its IoU with a ground truth instance is greater than $\tau_{iou}=0.5$. 

\paragraph{3D Detection.} We used a reference \textit{Car} size of $W_0=1.53m$, $H_0=1.63m$, $D_0=3.88m$ and depth statistics of $\mu_z=28.01m$ and $\sigma_z=16.32m$. We filtered the final 3D detections with a score threshold of $\tau_\text{conf}=0.05$. 

\paragraph{Losses.} We applied the same weighting policies in all our experiments. We set weight $1.0$ to all losses in the 2D detection head and $0.5$ to all losses in the 3D detection head.
The Huber parameters is set to $\delta_H=3.0$ and the 3D confidence temperature of $T=1$.

\paragraph{Optimization.} Our training schedule is the same for all experiments, and it does not involve any multi-step or warm-up procedures. We used SGD with a learning rate set at 0.01 and apply weight decay of 0.0001 to all parameters but scale and biases of \iABN. We also freeze conv1 and conv2 of ResNet34 in the backbone. We trained with batch size of 96 on 4 NVIDIA V-100 GPUs for a total of 20k iterations, scaling the learning rate by a 0.1 factor at 12k and 16k iterations. Our input resolution is set according to ~\cite{Manhardt_2019_CVPR}. We applied horizontal flipping as the only form of training-data augmentation. No augmentation was performed for test/validation.


\subsection{2D Detection}

In a first set of experiments, we study the signed IoU loss function (Sec.~\ref{sec:head2D}) in isolation.
To do this, we train our backbone + 2D head to perform pure 2D detection of cars in KITTI3D, comparing between the original RetinaNet regression loss, signed IoU and signed IoU with disentanglement.
For this simpler task we reduce the training schedule to 3.5k iterations, with learning rate steps after 2k and 3k, while keeping all other parameters as in Sec.~\ref{sec:details}.
As shown in Tab.~\ref{tab:results2d}, using signed IoU leads to a modest performance increase, which improves considerably when adding disentanglement.
\begin{table}[th]
    \centering
    {\footnotesize
    \begin{tabular}{l|ccc}
        \toprule
        Method & Easy & Moderate & Hard \\
        \midrule
        RetinaNet & 87.77 & 83.74 & 74.02 \\
        RetinaNet + IoU & 88.37 & 84.05 & 74.32 \\
        RetinaNet + \ioudis & \textbf{89.35} & \textbf{85.38} & \textbf{76.26} \\
        \bottomrule
    \end{tabular}}
    \caption{Ablation results on KITTI3D with 2D detection networks, $\text{AP}|_{R_{40}}$ scores.}
    \vspace{-1em}
    \label{tab:results2d}
\end{table}

\subsection{3D Detection}

In this section we focus on our main task and perform a detailed ablation of our contributions, comparing the results with most relevant state-of-the-art algorithms for monocular 3D detection.
Keeping the network architecture and training schedule fixed, we evaluate different loss functions and detection scoring strategies.
Following the discussion in Sec.~\ref{sec:metric}, we report both, our revised $\text{AP}|_{R_{40}}$ metric (Tab.~\ref{tab:ablation-new-metric}) and the original $\text{AP}|_{R_{11}}$ (Tab.~\ref{tab:ablation-old-metric}).


\paragraph{Ablation study.}
First, we turn our attention to the \textit{3D BB} loss in Eq.~\eqref{eq:det3d}, comparing it to the direct \textit{Regression} of the 10D parameters $\boldsymbol{\theta}$~\cite{Manhardt_2019_CVPR} (first two lines of both tables).
Confirming the findings in~\cite{Manhardt_2019_CVPR}, we observe increased 3D detection scores when tying all parameters together in a single (entangled) loss function in metric space.
Perhaps surprisingly, \textit{3D BB} also leads to better 2D detection performance: we suppose this could be due to more informative gradients propagating from the 3D head improving the backbone features.
Adding our disentangled 2D detection loss based on the signed IoU (Eq.~\eqref{eq:det2d}) and the 3D confidence prediction (Sec.~\ref{sec:head3D}), consistently improves performance for both \textit{Regression} and \textit{3D BB} (third and fourth lines in the tables).
Similarly, applying disentangling to the \textit{3D BB} loss improves 3D detection performance, and has an even larger impact on the 2D side.
Bringing all our contributions together leads to noticeable performance increases under all considered metrics (\textit{\monodis}).
In Tab.~\ref{tab:conf-2d} we conduct an additional ablation study on the validation set in~\cite{NIPS2015_Chen} to assess the importance of the 3D confidence prediction. To this end, we take our best model trained and evaluated with the 3D confidence prediction ($p_\mathtt{3D}, \text{AP}|_{R_{xx}}$) and compare against the same model when the 2D confidence is returned ($p_\mathtt{2D}, \text{AP}|_{R_{xx}}$) and when it is randomly sampled (random,~$\text{AP}|_{R_{xx}}$) . The ability of computing a reliable estimation of the confidence about the prediction is of utmost importance as can be inferred by the drastic drop of performance that we get when replacing $p_\mathtt{3D}$ with $p_\mathtt{2D}$, or with a random confidence. This is a direct consequence of the important role that the returned confidence plays in the AP metric.
\begin{table*}[th]
    \centering
    {\footnotesize
    \begin{tabular}{l|ccc|ccc|ccc}
        \toprule
        & \multicolumn{3}{c|}{2D detection} & \multicolumn{3}{c|}{3D detection} & \multicolumn{3}{c}{Bird's eye view} \\
        Method, metric & Easy & Moderate & Hard & Easy & Moderate & Hard & Easy & Moderate & Hard \\
        \midrule
	$p_\mathtt{3D}$, $\text{AP}|_{R_{11}}$ & 90.23 & 88.64 & 79.10 & 18.05 & 14.98 & 13.42 & 24.26 & 18.43 & 16.95 \\
	$p_\mathtt{2D}$, $\text{AP}|_{R_{11}}$ & 77.61 & 82.33 & 74.52 & 5.46 & 4.91 & 4.27 & 9.03 & 7.25 & 7.05 \\
	random, $\text{AP}|_{R_{11}}$ & 19.01 & 28.83 & 31.65 & 2.70 & 2.37 & 1.82 & 3.82 & 2.75 & 2.90 \\
        \midrule
	$p_\mathtt{3D}$, $\text{AP}|_{R_{40}}$ & 94.96 & 89.22 & 80.58 & 11.06 & 7.60 & 6.37 & 18.45 & 12.58 & 10.66 \\
	$p_\mathtt{2D}$, $\text{AP}|_{R_{40}}$ & 81.92 & 82.91 & 75.98 & 5.07 & 3.72 & 3.44 & 8.61 & 6.73 & 6.00 \\
	random, $\text{AP}|_{R_{11}}$ & 18.12 & 28.15 & 29.88 & 1.72 & 1.51 & 1.36 & 2.96 & 2.42 & 2.58 \\
        \bottomrule
    \end{tabular}}
    \caption{Results on KITTI3D when using $p_\mathtt{2D}$ or $p_\mathtt{3D}=p_\mathtt{3D|2D}p_\mathtt{2D}$ as the final confidence score to rank predictions. In addition, we report the performance when the confidence is sampled from a uniform distribution.}
    \label{tab:conf-2d}
    \vspace{-12pt}
\end{table*}

\paragraph{Comparison with SOTA.}
In Tab.~\ref{tab:ablation-new-metric}, \ref{tab:ap11-test} and \ref{tab:ablation-old-metric} we report validation and test set results, respectively, of many recent monocular 3D detection approaches.
When evaluating on the validation set, we consider the split defined in~\cite{NIPS2015_Chen}, as is done in all the baselines. Please note that the works in~\cite{Xiang_2015_CVPR,Mousavian_2017_CVPR,Xiang_2017_WACV} are using yet another training/validation split, rendering their results incomparable to ours while yielding numerically comparable ranges to \eg~\cite{Liu+19}.
For the test set, we consider both the split in~\cite{NIPS2015_Chen}, which is shared with OFTNet~\cite{Roddick18} and ROI-10D~\cite{Manhardt_2019_CVPR}, and a larger training split\footnote{\url{https://github.com/MarvinTeichmann/KittiBox}}, since the setting used for MonoGRNet~\cite{qin2019monogrnet} is not clear.
In Tab.~\ref{tab:ablation-new-metric} we show $\text{AP}|_{R_{40}}$ scores\footnote{We calculated these from the precision-recall values published in the KITTI3D leaderboard page.} for the test set results, and in Tb.~\ref{tab:ap11-test} the corresponding $\text{AP}|_{R_{11}}$ scores.
Nonetheless, we would like to stress that the $\text{AP}|_{R_{11}}$ is biased by the issue reported in Section~\ref{sec:metric} and we invite to rather consider $\text{AP}|_{R_{40}}$ as the reference metric for fair comparison.
With a single exception, our approach beats all baselines on all 3D and bird's eye view metrics, often by a large margin, despite the fact that
some of the outperformed methods rely on additional data, such as synthetic images (ROI-10D~\cite{Manhardt_2019_CVPR}), or a pre-trained monocular depth prediction network (ROI-10D~\cite{Manhardt_2019_CVPR}, Xu~\etal~\cite{Xu_2018_CVPR}).
Interestingly, from the validation set results in Tab.~\ref{tab:ablation-old-metric}, many existing approaches score lower than the ``single correct hypothesis'' baseline (see Sec.~\ref{sec:metric}) on 3D detection $\text{AP}|_{R_{11}}$, highlighting the need for an improved AP metric.
\newcommand{\gc}{0.9}
\begin{table*}[t]
    \centering
    {\footnotesize
    \begin{tabular}{l|ccc|ccc|ccc}
        \toprule
        & \multicolumn{3}{c|}{2D detection} & \multicolumn{3}{c|}{3D detection} & \multicolumn{3}{c}{Bird's eye view} \\
        Method & Easy & Moderate & Hard & Easy & Moderate & Hard & Easy & Moderate & Hard \\
        \midrule
        Regression & 70.10 & 73.20 & 66.80 & 1.30 & 0.90 & 0.70 & 2.60 & 1.90 & 1.70 \\
        3D BB & 74.30 & 77.10 & 69.50 & 3.90 & 2.70 & 2.50 & 6.90 & 5.10 & 4.40 \\
        \midrule
        Regression w/ \ioudis, \confidence & 70.10 & 75.10 & 66.90 & 2.60 & 1.70 & 1.40 & 5.40 & 3.80 & 3.00 \\
        3D BB w/ \ioudis, \confidence & \textbf{95.10} & 88.90 & 78.60 & 8.80 & 6.10 & 5.00 & 14.60 & 10.10 & 8.30 \\
        3D BB w/ disentangling & 80.50 & 80.80 & 74.40 & 4.10 & 3.00 & 2.70 & 7.10 & 5.40 & 4.80 \\
        \rowcolor{mapillarygreen}
        \monodis & 94.96 & \textbf{89.22} & \textbf{80.58} & \textbf{11.06} & \textbf{7.60} & \textbf{6.37} & \textbf{18.45} & \textbf{12.58} & \textbf{10.66} \\
        \midrule
        \rowcolor[gray]{\gc}
        OFTNet~\cite{Roddick18} & -- & -- & -- & 1.61 & 1.32 & 1.00 & 1.28 & 0.81 & 0.51 \\
        \rowcolor[gray]{\gc}
        FQNet~\cite{Liu+19} & \textbf{94.72} & \textbf{90.17} & 76.78 & 2.77 & 1.51 & 1.01 & 5.40 & 3.23 & 2.46 \\
        \rowcolor[gray]{\gc}
        ROI-10D w/ Depth, Synthetic~\cite{Manhardt_2019_CVPR} & 76.56 & 70.16 & 61.15 & 4.32 & 2.02 & 1.46 & 9.78 & 4.91 & 3.74 \\
        \rowcolor[gray]{\gc}
        MonoGRNet~\cite{qin2019monogrnet} & 88.65 & 77.94 & 63.31 & 9.61 & 5.74 & 4.25 & \textbf{18.19} & 11.17 & 8.73 \\
        \rowcolor{mapillarygreen}
        \monodis{} & 93.11 & 85.86 & 73.61 & 7.03 & 4.89 & 4.08 & 12.18 & 9.13 & 7.38 \\
        \rowcolor{mapillarygreen}
        \monodis{}, larger training split
        & 94.61 & 89.15 & \textbf{78.37} & \textbf{10.37} & \textbf{7.94} & \textbf{6.40} & 17.23 & \textbf{13.19} & \textbf{11.12} \\
        \bottomrule 
    \end{tabular}}
    \vspace{-0.5em}
    \caption{$\text{AP}|_{R_{40}}$ scores on KITTI3D: ablation results (white background), test set results of SOTA (grey background) and ours (green background).}
    \label{tab:ablation-new-metric}
    \vspace{-0.5em}
\end{table*}
\begin{table*}
    \centering
    {\footnotesize
    \begin{tabular}{l|ccc|ccc|ccc}
        \toprule
        & \multicolumn{3}{c|}{2D detection} & \multicolumn{3}{c|}{3D detection} & \multicolumn{3}{c}{Bird's eye view} \\
        Method & Easy & Moderate & Hard & Easy & Moderate & Hard & Easy & Moderate & Hard \\
        \midrule
        \rowcolor[gray]{\gc}
        OFTNet~\cite{Roddick18} & -- & -- & -- & 3.28 & 2.50 & 2.27 & 9.50 & 7.99 & 7.51 \\
        \rowcolor[gray]{\gc}
        FQNet~\cite{Liu+19} & \textbf{90.45} & \textbf{88.83} & \textbf{77.55} & 3.48 & 2.42 & 1.96 & 6.51 & 4.62 & 3.99 \\
        \rowcolor[gray]{\gc}
        ROI-10D w/ Depth, Synthetic~\cite{Manhardt_2019_CVPR} & 75.33 & 69.64 & 61.18 & \textbf{12.30} & 10.30 & 9.39 & 16.77 & 12.40 & 11.39 \\
        \rowcolor[gray]{\gc}
        MonoGRNet~\cite{qin2019monogrnet} & 87.23 & 77.46 & 61.12 & 11.29 & 12.90 & 11.34 & \textbf{20.55} & 16.37 & 15.16 \\
        \rowcolor{mapillarygreen}
        \monodis{} & 89.61 & 83.80 & 70.84 & 8.26 & 6.15 & 6.06 & 13.10 & 11.12 & 9.35 \\
        \rowcolor{mapillarygreen}
        \monodis{}, larger training split
        & 90.31 &  87.58 & 76.85 & 11.81 & \textbf{15.12}& \textbf{12.71} & 18.88 & \textbf{19.08} & \textbf{17.41} \\
        \bottomrule
    \end{tabular}}
    \caption{$\text{AP}|_{R_{11}}$ scores on KITTI3D: test set results of SOTA (grey background) and ours (green background).}
    \label{tab:ap11-test}
    \vspace{-10pt}
\end{table*}

\begin{table*}[t]
    \centering
    {\footnotesize
    \begin{tabular}{l|ccc|ccc|ccc}
        \toprule
        & \multicolumn{3}{c|}{2D detection} & \multicolumn{3}{c|}{3D detection} & \multicolumn{3}{c}{Bird's eye view} \\
        Method & Easy & Moderate & Hard & Easy & Moderate & Hard & Easy & Moderate & Hard \\
        \midrule
        Regression & 66.50 & 72.30 & 66.00 & 1.60 & 1.50 & 1.20 & 2.70 & 2.10 & 2.30 \\
        3D BB & 70.80 & 77.10 & 66.50 & 4.70 & 3.00 & 2.90 & 7.80 & 5.40 & 5.80 \\
        \midrule
        Regression w/ \ioudis, \confidence & 67.20 & 73.60 & 65.50 & 3.20 & 2.90 & 2.00 & 5.80 & 4.80 & 4.30 \\
        3D BB w/ \ioudis, \confidence & 90.20 & 88.40 & 78.40 & 15.40 & 13.60 & 12.00 & 20.50 & 16.20 & 15.70 \\
        3D BB w/ disentangling & 76.40 & 80.30 & 73.20 & 4.90 & 3.40 & 3.10 & 7.30 & 5.70 & 6.30 \\
        \rowcolor{mapillarygreen}
        \monodis{} & 90.23 & 88.64 & 79.10 & \textbf{18.05} & \textbf{14.98} & \textbf{13.42} & \textbf{24.26} & \textbf{18.43} & \textbf{16.95} \\
        \midrule
        \rowcolor[gray]{\gc}
        Single correct hypothesis per difficulty  & 9.09 & 9.09 & 9.09 & 9.09 & 9.09 & 9.09 & 9.09 & 9.09 & 9.09 \\
        \rowcolor[gray]{\gc}
        OFTNet~\cite{Roddick18} & -- & -- & -- & 4.07 & 3.27 & 3.29 & 11.06 & 8.79 & 8.91 \\
        \rowcolor[gray]{\gc}
        Xu~\etal~\cite{Xu_2018_CVPR} & -- & -- & -- & 7.85 & 5.39 & 4.73 & 19.20 & 12.17 & 10.89 \\
        \rowcolor[gray]{\gc} 
        FQNet~\cite{Liu+19} & -- & -- & -- & 5.98 & 5.50 & 4.75 & 9.50 & 8.02 & 7.71 \\
        \rowcolor[gray]{\gc}
        Mono3D~\cite{Chen_2016_CVPR} & \textbf{93.89} & \textbf{88.67} & \textbf{79.68} & 2.53 & 2.31 & 2.31 & 5.22 & 5.19 & 4.13 \\
        \rowcolor[gray]{\gc}
        Mono3D++~\cite{TongHe_2019_arxiv} & -- & -- & -- & 10.60 & 7.90 & 5.70 & 16.70 & 11.50 & 10.10 \\
        \rowcolor[gray]{\gc}
        ROI-10D~\cite{Manhardt_2019_CVPR} & 78.57 & 73.44 & 63.69 & 10.12 & 1.76 & 1.30 & 14.04 & 3.69 & 3.56 \\
        \rowcolor[gray]{\gc}
        ROI-10D w/ Depth~\cite{Manhardt_2019_CVPR} & 89.04 & 88.39 & 78.77 & 7.79 & 5.16 & 3.95 & 10.74 & 7.46 & 7.06 \\
        \rowcolor[gray]{\gc}
        ROI-10D w/ Depth, Synthetic~\cite{Manhardt_2019_CVPR} & 85.32 & 77.32 & 69.70 & 9.61 & 6.63 & 6.29 & 14.50 & 9.91 & 8.73 \\
        \rowcolor[gray]{\gc}
        MonoGRNet~\cite{qin2019monogrnet} & -- & -- & -- & 13.88 & 10.19 & 7.62 & -- & -- & -- \\
        \rowcolor[gray]{\gc}
        Best in~\cite{Barabanau_arXiv_2019} & -- & -- & -- & 13.96 & 7.37 & 4.54 & -- & -- & -- \\
        \bottomrule
    \end{tabular}}
    \vspace{-0.5em}
    \caption{$\text{AP}|_{R_{11}}$ scores on KITTI3D (0.7 IoU threshold): Ablation results (white background), val set results of SOTA (grey background).}
    \label{tab:ablation-old-metric}
    \vspace{-1em}
\end{table*}

\paragraph{Results on additional KITTI3D classes.}
In Tab.~\ref{tab:ped_cyc} we provide the $\text{AP}|_{R_{11}}$ and $\text{AP}|_{R_{40}}$ scores (at IoU treshold $0.5$, see official evaluation scripts) obtained on the validation set in~\cite{NIPS2015_Chen} for classes \textit{Pedestrian} and \textit{Cyclist} (trained independently). If compared to the results on class \textit{Car}, it can be seen that performances on these two particular classes are in general lower. The performance degradation on classes \textit{Pedestrian} and \textit{Cyclist} compared to \textit{Car} is due to i) the reduced number of annotations which is $\approx6\times$ and $\approx20\times$ lower than \emph{Car} for class \textit{Pedestrian} and \textit{Cyclist}, respectively, and ii) the higher impact that errors on localization have on the AP scores since the object $xz$-extent is typically smaller. For these reasons, similarly to~\cite{Manhardt_2019_CVPR,qin2019monogrnet,Roddick18,Xu_2018_CVPR}, we put a larger focus on class \textit{Car} in the main paper. 
\begin{table*}[ht]
    \centering
    {\footnotesize
    \begin{tabular}{l|ccc|ccc|ccc}
        \toprule
        & \multicolumn{3}{c|}{2D detection} & \multicolumn{3}{c|}{3D detection} & \multicolumn{3}{c}{Bird's eye view} \\
        Method, metric, class & Easy & Moderate & Hard & Easy & Moderate & Hard & Easy & Moderate & Hard \\
        \midrule
	\monodis{}, $\text{AP}|_{R_{11}}$, pedestrian & 72.16 & 64.93 & 56.89 & 10.79 & 10.39 & 9.22 & 11.04 & 10.94 & 10.59 \\
	\monodis{}, $\text{AP}|_{R_{40}}$, pedestrian & 72.78 & 65.56 & 56.50 & 3.20 & 2.28 & 1.71 & 4.04 & 3.19 & 2.45 \\
        \midrule
	\monodis{}, $\text{AP}|_{R_{11}}$, cyclist & 67.81 & 49.15 & 47.26 & 5.27 & 4.55 & 4.55 & 5.52 & 4.66 & 4.55 \\
	\monodis{}, $\text{AP}|_{R_{40}}$, cyclist & 68.12 & 47.45 & 45.60 & 1.52 & 0.73 & 0.71 & 1.87 & 1.00 & 0.94 \\
        \bottomrule
    \end{tabular}}
    \caption{Results on the classes \textit{Pedestrian} and \textit{Cyclist} on the KITTI3D validation set (0.5 IoU threshold).}
    \label{tab:ped_cyc}
    \vspace{-12pt}
\end{table*}

\paragraph{Qualitative results.} In Fig.~\ref{fig:car_viz} we show qualitative results on a set of images taken from the validation set for the classes \textit{Car} (top), \textit{Pedestrian} (middle) and \textit{Cyclist} (bottom).
We also provide a video\footnote{\url{https://research.mapillary.com/publication/MonoDIS}} showing detection results obtained on a sequence from the validation set. The structure of the frames is similar to the one in Fig.~\ref{fig:car_viz}, where detections are shown on the right side and the corresponding birds-eye view on the left. For simplicity, we decided to display all the detections with the same color.



\section{Experiments on nuScenes}

We conduct additional experiments on the novel \textit{nuScenes} dataset~\cite{Cae+19}.

\paragraph{About the dataset.}
The nuScenes dataset provides multimodal, street-level data collected with a car equipped with 6 cameras, 1 LiDAR, 5 Radars and IMU. It contains 15h of driving data (242 km at average speed of 16 km/h) covering parts of the areas of Boston (Seaport and South Boston) and Singapore (One North, Holland Village and Queenstown). These two cities have been chosen due to their known dense traffic and highly challenging driving situations and driving routes are selected to capture a diverse set of locations, times and weather conditions.
The dataset provides $360^\circ$, synchronized sensor coverage, calibration of sensor intrinsics and extrinsics parameters, and objects annotations for $23$ different classes from $1000$ selected scenes of $20s$ duration each. Annotated objects in the scenes come with a semantic category, 3D bounding box, tracking information, and attributes (visibility, activity and pose) for each frame they occur in.

\paragraph{Detection task.}
The nuScenes detection tasks requires detecting $10$ object classes in terms of full 3D bounding boxes, attributes and velocities. 
In this work, we will focus on detecting the full 3D bounding box of object belonging to class \emph{car}, because the only available baselines at the time of writing are OFTNet (monocular RGB image-based) and PointPillar~\cite{Lang_CVPR_2019} (LiDAR-based). Fair comparison can only be made to OFTNet, where results are reported only for category \textit{car} (see, \cite{Cae+19}).

\paragraph{Evaluation metric.}
The authors of nuScenes propose an alternative metric called \emph{nuScenes detection score} (NDS) that combines a measure of the detection performance with quality terms of box location (ATE, average translation error), size (ASE, average scale error), orientation (AOE, average orientation error), attributes (AAE, average attribute error) and velocity (AVE, average velocity error). The detection performance is measured in terms of Average Precision (AP), but with matches determined based on 2D center distance on the ground plane. Also the AP score is calculated as the normalized area under the precision/recall curve by excluding the $[0-10\%]$ range. The final score averages AP over matching thresholds of $\mathbb D=\{0.5,1,2,4\}$ meters and the set of classes $\mathbb C$:
\[
	\text{mAP}=\frac{1}{|\mathbb C| |\mathbb D|}\sum_{c\in\mathbb C}\sum_{d\in\mathbb D} \text{AP}_{c,d}\,,
\]
where $\text{AP}_{c,d}$ is the AP score on class $c$ with matching threshold $d$. 

\begin{figure}
    \centering
    \includegraphics[width=0.51\columnwidth]{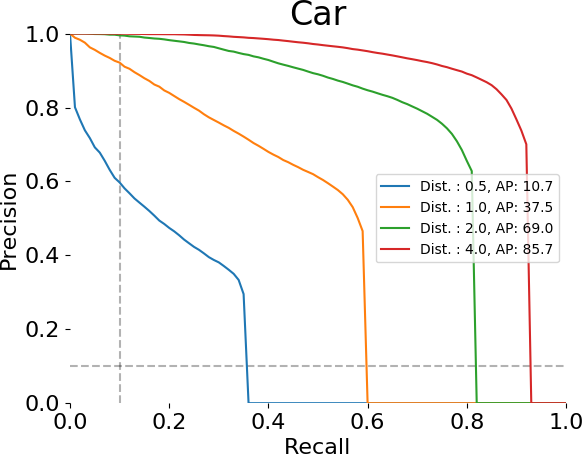}\hfill
    \includegraphics[width=0.47\columnwidth]{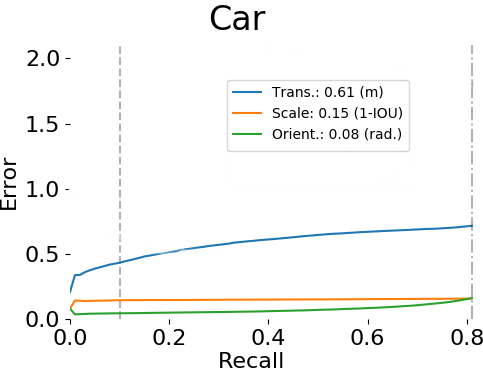}
    \caption{Performance plots for class \textit{Car} in nuScenes. Left: Precision/Recall curves for AP metric at multiple distance thresholds in $\mathbb D$. Right: Error/Recall curves for relevant TP errors metrics on translation (ATE), scale (ASE) and orientation (AOE).}
    \label{fig:nuscenesRes}
    \vspace{-12pt}
\end{figure}

\paragraph{Obtained results.}
We present in Fig.~\ref{fig:nuscenesRes} the results obtained on the car class in terms of Precision/Recall curves (for all distance thresholds in $\mathbb D$), as well as error curves for translation, scale and orientation true positive metrics (at distance threshold 2m), produced by the official nuScenes evaluation scripts. For direct comparison to available OFTNet and PointPillar results from~\cite{Cae+19}, we also provide Tab.~\ref{tab:nuscenesTab}. It is important to stress that direct comparison is only fair to OFTNet which is also purely image-based, unlike PointPillar, which is LiDAR-based. We are not reporting the NDS score as it also requires predictions for attributes and velocities. Since that would imply modifications of the network design it would also render results inconsistent with those obtained on KITTI3D in Sec.~\ref{sec:expKITTI}. 

The results in Tab.~\ref{tab:nuscenesTab} show that our approach improves by \textbf{42\%} over OFTNet (in absolute terms), considering the primary AP metric at a distance threshold of 2m. In addition, \monodis{} improves on all available True Positive metrics over OFTNet and even on 2/3 metrics when compared to PointPillar (LiDAR-based). Despite obtaining better (lower) TP metrics ASE and AOE compared to PointPillar, the main advantage of LiDAR-based methods are shown in their lower translation errors (and therefore also in the corresponding AP scores at various distances). We provide some qualitative results in Fig.~\ref{fig:nuscenesViz}, demonstrating promising 3D recognition performance without using LiDAR and therefore actively sensed depth information.

\begin{table}[th]
    \centering
    \resizebox{\columnwidth}{!}{
    \begin{tabular}{l|cccc|ccc}
        \toprule
         & \multicolumn{4}{c}{AP$_{\mbox{Car}}\uparrow$ [\%]}  & \multicolumn{3}{c}{TP$_{\mbox{Car}}\downarrow$} \\
         Method & 0.5m & 1.0m & \textbf{2.0m} & 4.0m & ATE [m] & ASE [1-IoU] & AOE [rad] \\
        \midrule
        PointPillar &  55.5 & 71.8 & 76.1 & 78.6 &  0.27 & 0.17 & 0.19 \\
        \midrule
        \rowcolor[gray]{\gc}
        OFTNet  &  -- & -- & 27.0 & -- & 0.65 & 0.16 & 0.18 \\
        \rowcolor{mapillarygreen}
        MonoDIS & 10.7 & 37.5 & \textbf{69.0} & 85.7 & \textbf{0.61} & \textbf{0.15} & \textbf{0.08} \\        
        \bottomrule
    \end{tabular}}
    \caption{Performance comparison for results on category car in nuScenes dataset~\cite{Cae+19}. Top row: LiDAR-based PointPillar results (listed for completeness). Bottom: Available OFTNet results vs.~\monodis{}.} 
    \vspace{-1em}
    \label{tab:nuscenesTab}
\end{table}

\section{Conclusions}
We proposed a new loss disentangling transformation that allowed us to effectively train a 3D object detection network end-to-end without the need of stage-wise training or warm-up phases. Our solution isolates the contribution made by groups of parameters to a given loss into separate terms that retain the same nature of the original loss, thus being compatible without the need of further, cumbersome loss balancing steps. We proposed two further loss functions where i) is based on a novel signed Intersection-over-Union criterion to improve 2D detection results and ii) is used to predict a detection confidence for the 3D bounding box predictions, learned in a self-supervised way. Besides the methodological contributions, we reveal a flaw in the primary detection metric used in KITTI3D, where a single, correctly predicted bounding box yields overall AP scores of 9.09\% on validation or test splits. Our simple fix corrects performance results of previously published methods in general, and shows how significantly it was biasing monocular 3D object detection results in particular. In our extensive experimental results and ablation studies we demonstrated the effectiveness of our proposed model, and significantly improved over previous state-of-the-art on both, KITTI3D and the novel nuScenes dataset.

\appendix

\section{Signed Intersection over Union}\label{sec:siou}
\begin{figure}[ht]
    \centering
    \includegraphics[width=\columnwidth]{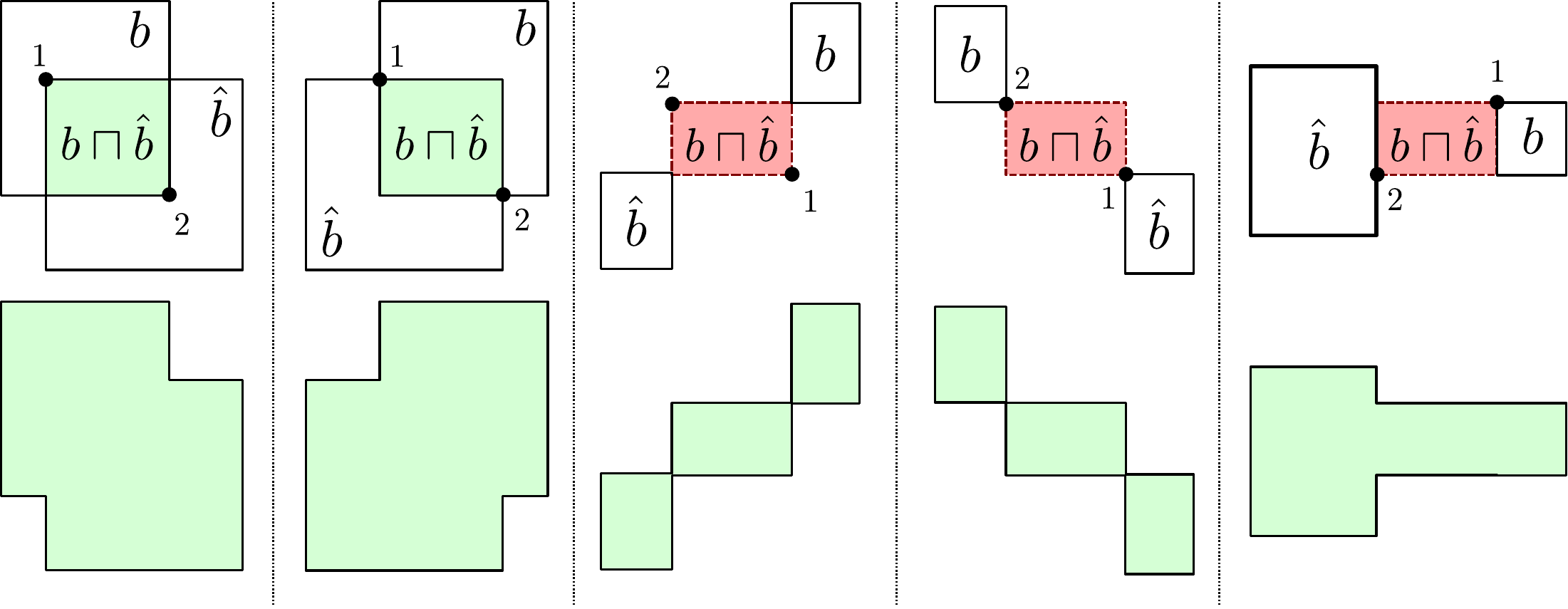}
    \caption{Five examples of computation of the proposed signed IoU. Top: Colored areas represent the numerator of the sIoU formula, where green denotes positive area, red denotes negative area; numbers represent the corner ordering. Bottom: Areas represent the denominator, which is always positive. }
    \label{fig:sIoU}
\end{figure}

Let $\vct{\hat b}=(\hat u_1,\hat v_1,\hat u_2,\hat v_2)$ and $\vct b=(u_1,v_1,u_2,v_2)$ be two bounding boxes, where $(u_1,v_1)$ denotes the top-left corner and $(u_2, v_2$) denotes the bottom-right corner.
We define the signed intersection-over-union as follows:
\begin{equation}\label{eq:sIoU}
\text{sIoU}(\vct b,\vct{\hat b})=\frac{|\vct b\sqcap\vct{\hat b}|_\pm}{|\vct b|+|\vct {\hat b}|-|\vct b\sqcap\vct{\hat b}|_\pm}\,,
\end{equation}
where
\[
\vct b\sqcap\vct{\hat b}=
\begin{pmatrix}
\max(u_1,\hat u_1)\\
\max(v_1,\hat v_1)\\
\min(u_2,\hat u_2)\\
\min(v_2,\hat v_2)
\end{pmatrix}
\]
provides an extended intersection operation between bounding boxes, $|\vct b|$ gives the area of bounding box $\vct b$ and
\[
|\vct b|_\pm=
\begin{cases}
+|\vct b|&\text{if } u_2>u_1\text{ and }v_2>v_1,\\
-|\vct b|&\text{otherwise,}
\end{cases}
\]
gives the \emph{signed} area of $\vct b$, which corresponds to the standard area with positive sign only if the first corner of $\vct b$ is the top-left one, while the second corner is the bottom-right one. To give a better intuition we provide some examples in Fig.~\ref{fig:sIoU}, where green and red colors encode positive and negative areas, respectively: Left-to-right, the first two examples boil down to standard IoU yielding positive values, while the last ones are examples yielding negative values. The sIoU score is bounded in $[-1,1]$.

\begin{figure*}[p]
    \centering
    \includegraphics[width=.8\textwidth]{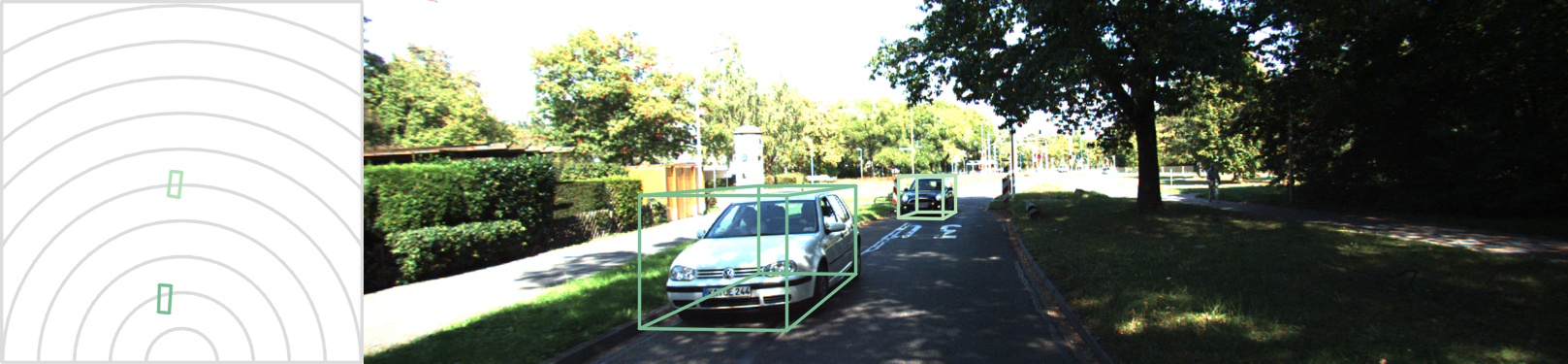}
    \includegraphics[width=.8\textwidth]{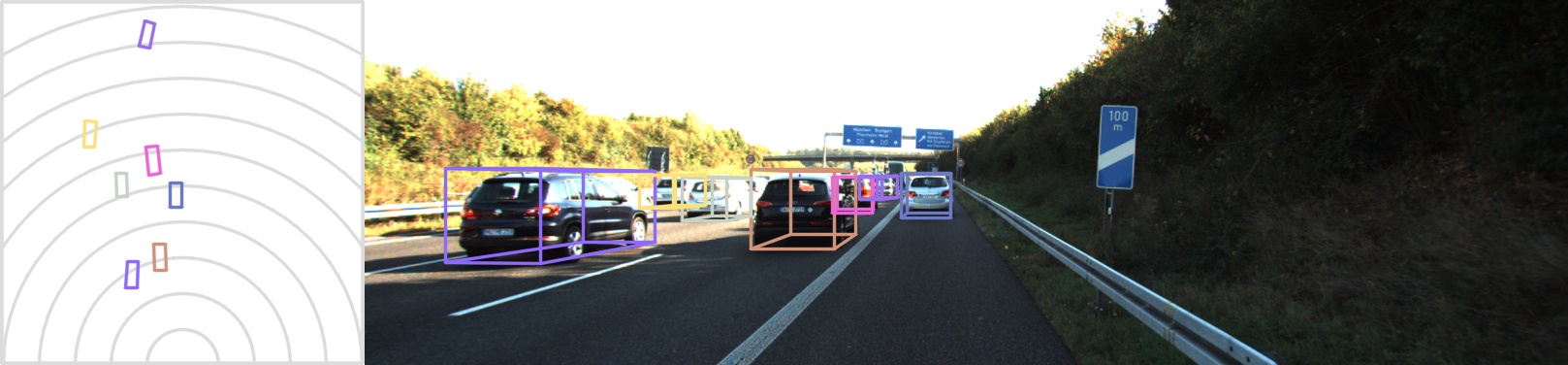}

    \vspace{15pt}
    \includegraphics[width=.8\textwidth]{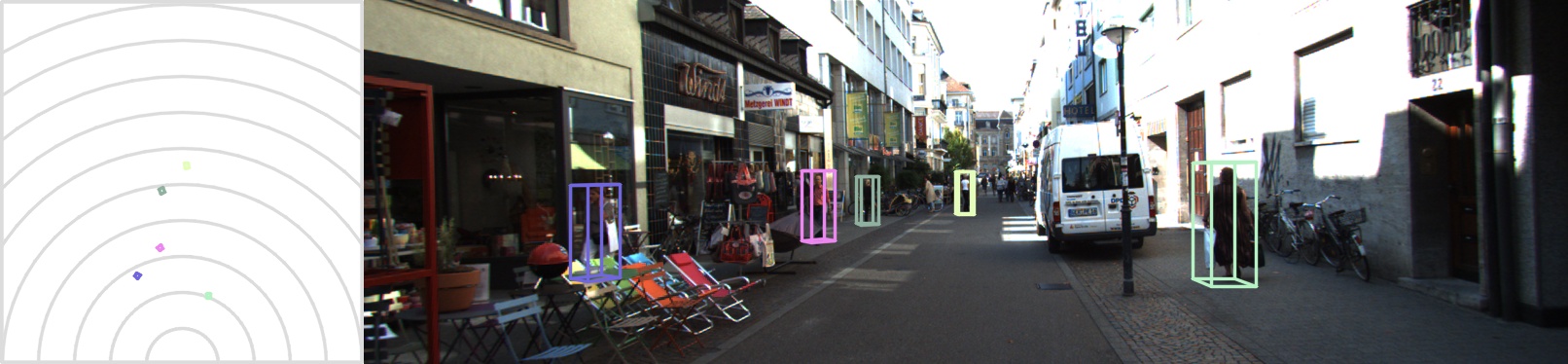}
    \includegraphics[width=.8\textwidth]{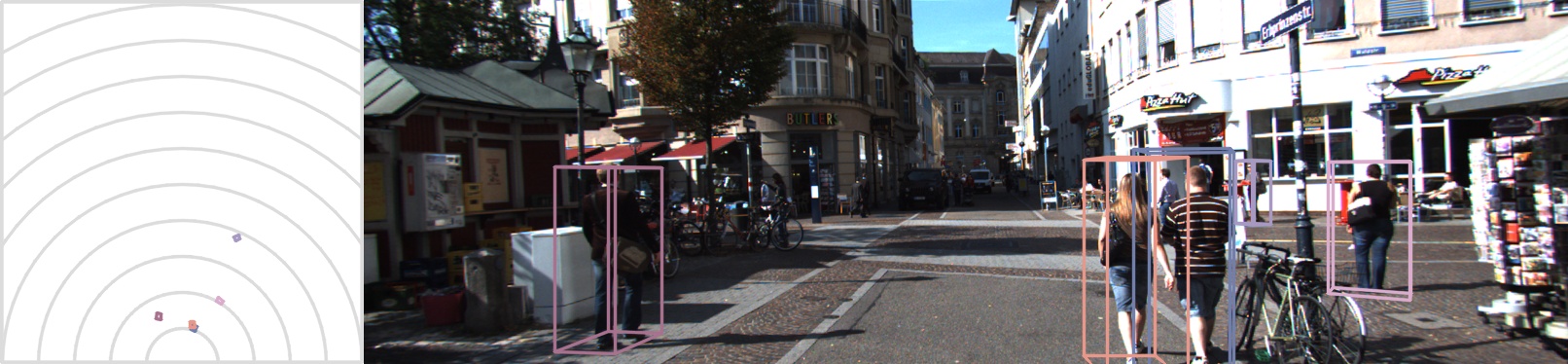}

    \vspace{15pt}
    \includegraphics[width=.8\textwidth]{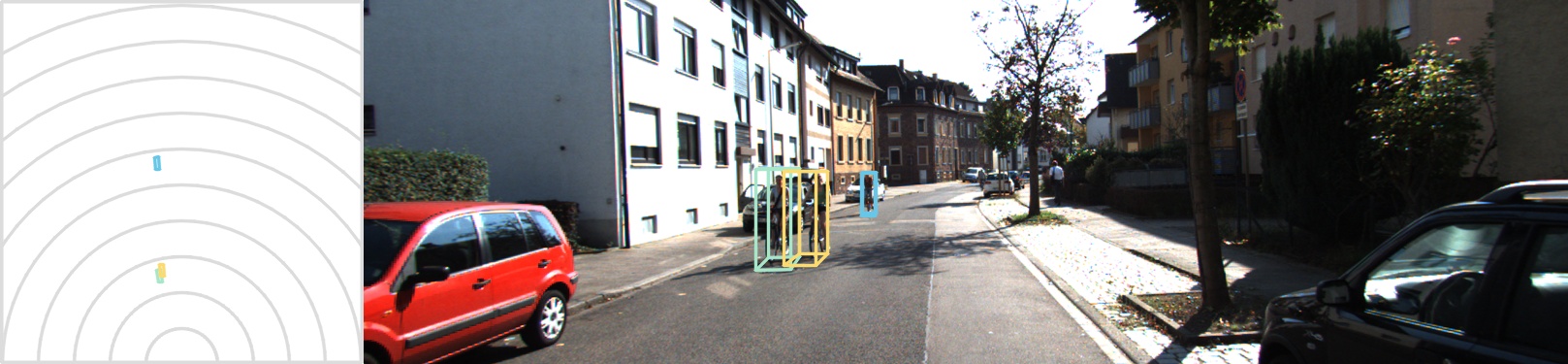}
    \includegraphics[width=.8\textwidth]{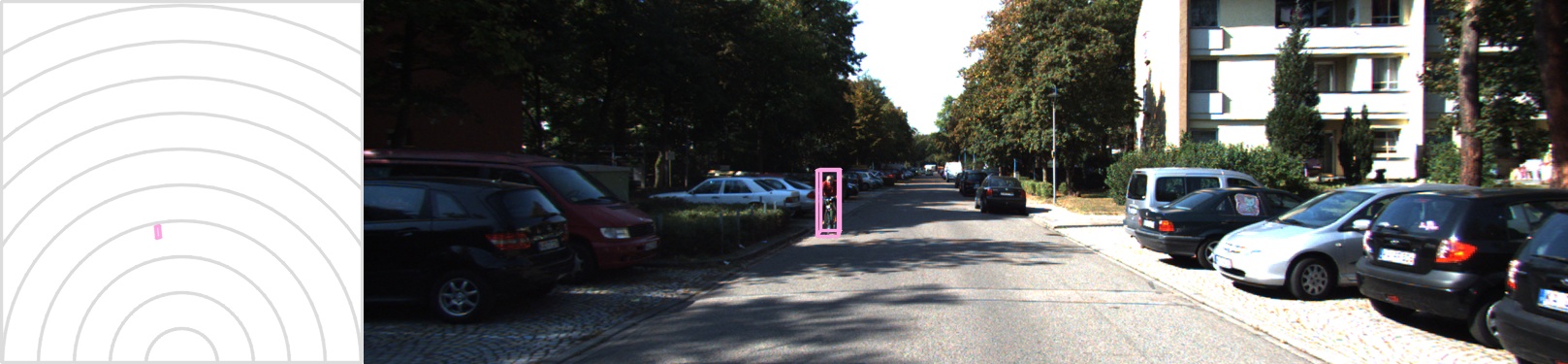}
    \caption{Example results for classes \textit{Car} (top), \textit{Pedestrian} (middle) and \textit{Cyclist}(bottom) with corresponding birds-eye view.}
    \label{fig:car_viz}
\end{figure*}

\begin{figure*}[p]
    \centering
    \includegraphics[width=.48\textwidth]{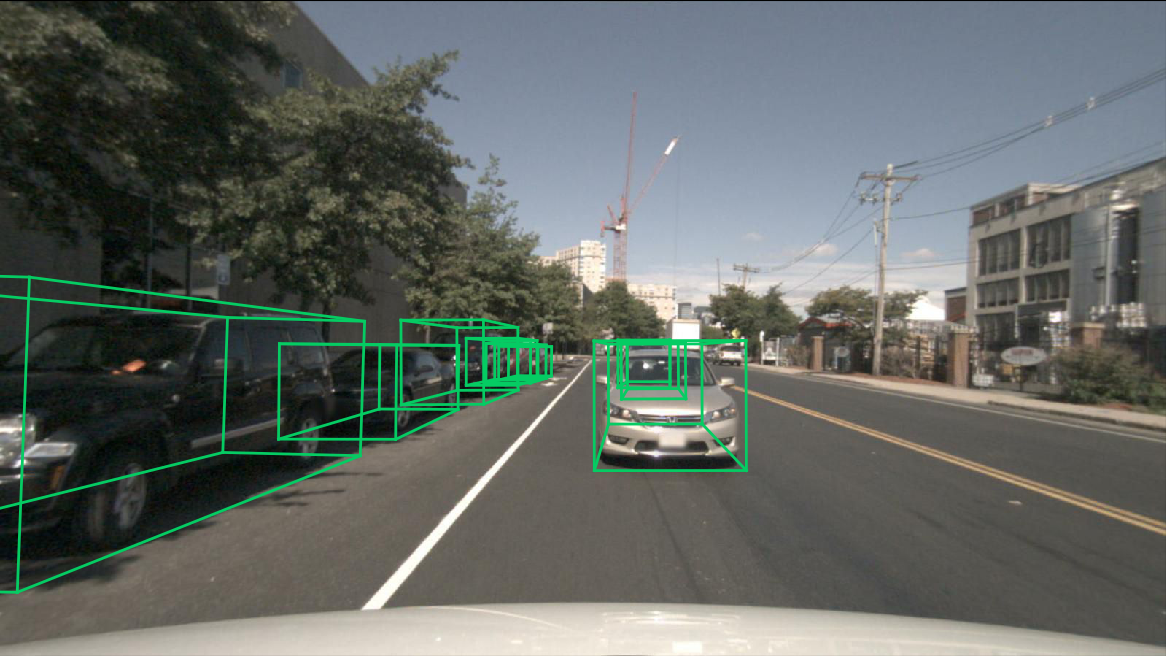}\hfill
    \includegraphics[width=.48\textwidth]{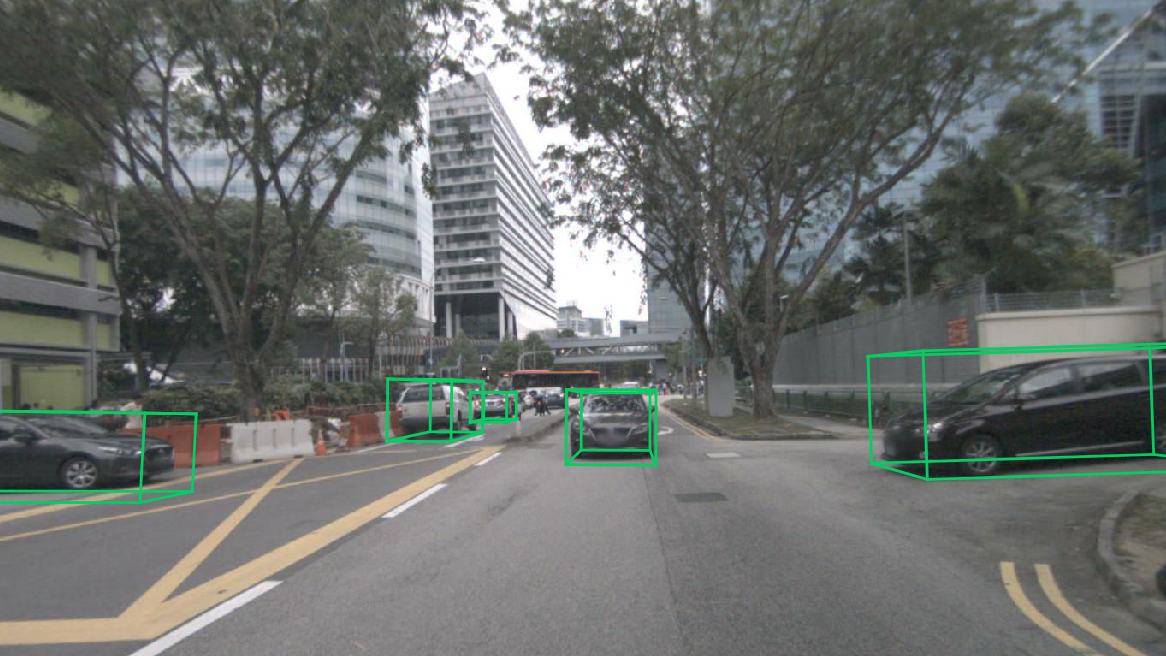}

    \vspace{15pt}
    \includegraphics[width=.48\textwidth]{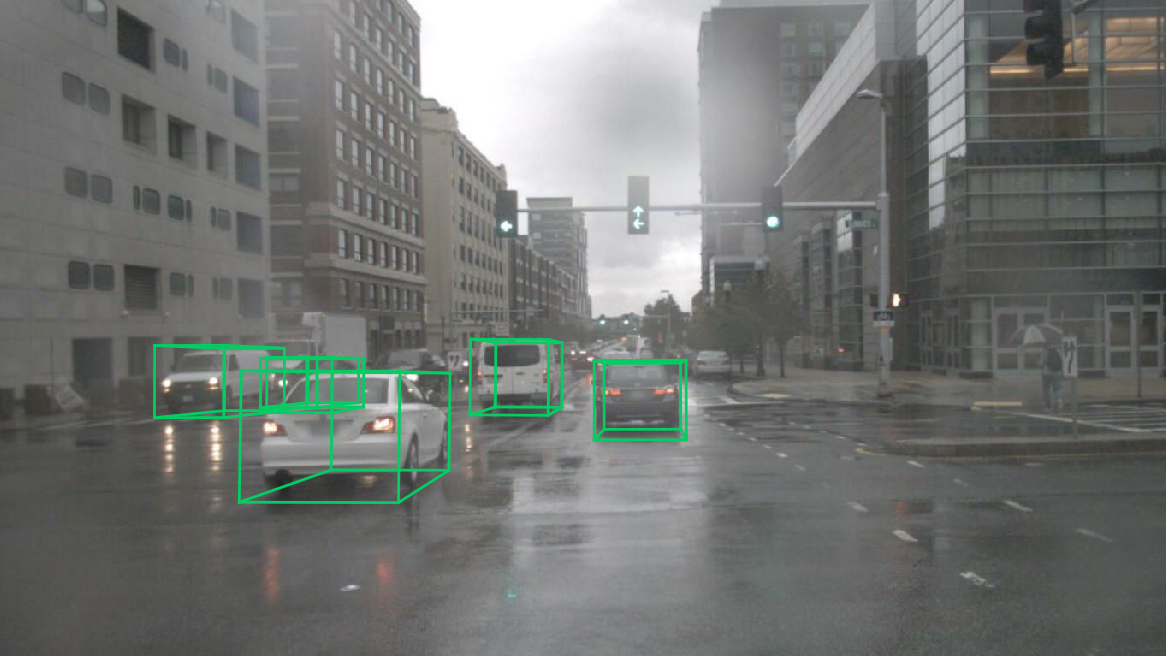}\hfill
    \includegraphics[width=.48\textwidth]{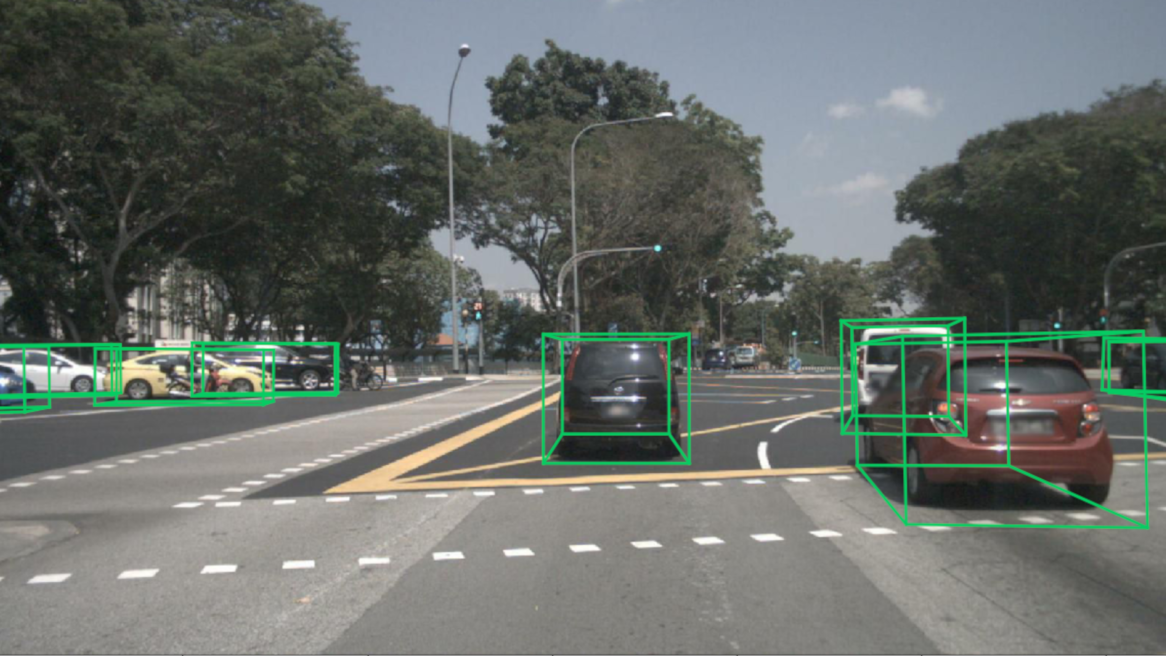}

    \vspace{15pt}
    \includegraphics[width=.48\textwidth]{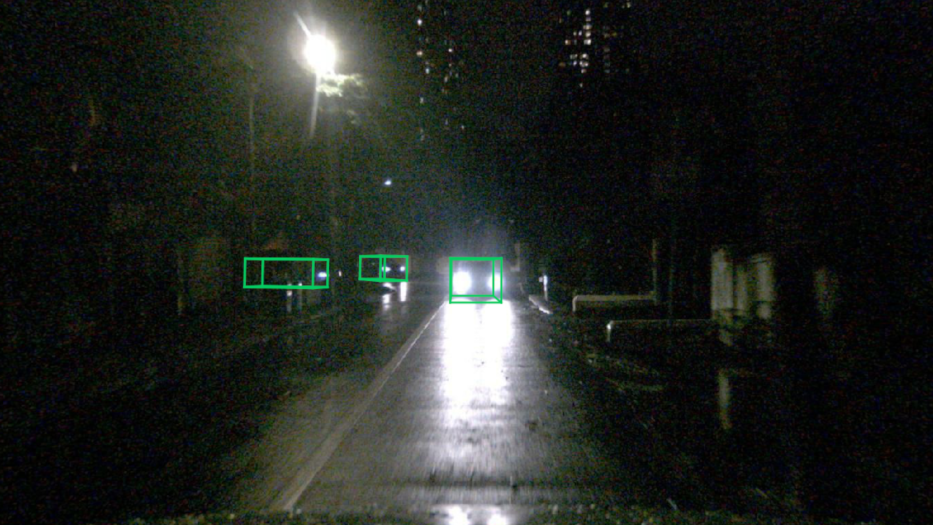}\hfill
    \includegraphics[width=.48\textwidth]{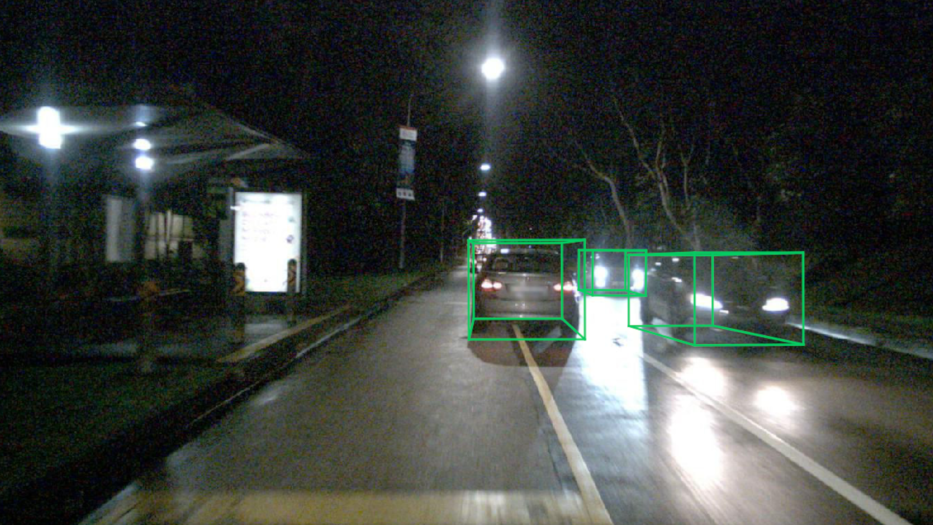}
    \caption{Example results for class \textit{Car} on nuScenes dataset for images taken at different weather and illumination conditions.}
    \label{fig:nuscenesViz}
\end{figure*}

\section{Lifting Transformation}\label{sec:lifting}
We review the lifting transformation used in~\cite{Manhardt_2019_CVPR}.
Let $\vct \theta$ be the 10D network's output from which we compute the depth $z$ of the 3D bounding box's center, its projection on the image place $\vct c=(u_c,v_c)$, the dimensions of the 3D bounding box $\vct s=(W,H,D)$ and the unit quaternion $\vct q$ as described in Sec.~4.3 of the main paper. Let $K$ be the $3\times 3$ matrix of intrinsics with entries:
\[
K=
\begin{bmatrix}
f_x&0&c_x\\
0&f_y&c_y\\
0&0&1
\end{bmatrix}
\]
and let
\[
\vct C=
\begin{pmatrix}
\frac{u_c-c_x}{f_x}z,&
\frac{v_c-c_y}{f_y}z,&
z
\end{pmatrix}^\top=(C_x,C_y,C_z)\T
\]
be the position of the center of the 3D bounding box.
The lifting transformation is defined as:
\[
\set F(\vct \theta)= \frac{1}{2}R_{\vct q_{\vct c}}S\,B_0+\vct C
\]
where $B_0$ holds the corners of the unit cube $[-1,1]^3$, $S$ is the diagonal matrix with entries $\vct s$, and $R_{\vct q_c}$ is the $3\times 3$ rotation matrix corresponding to quaternion 
\[
\vct q_c=\vct q 
\left[\cos \frac{\beta}{2} +sin \frac{\beta}{2} \mathtt j\right]
\]
with $\beta=\tan^{-1}(\frac{C_x}{C_z})$.

{
\small


\begin{thebibliography}{10}\itemsep=-1pt

\bibitem{Barabanau_arXiv_2019}
I.~Barabanau, A.~Artemov, E.~Burnaev, and V.~Murashkin.
\newblock Monocular 3d object detection via geometric reasoning on keypoints.
\newblock {\em CoRR}, abs/1905.05618, 2019.

\bibitem{Cae+19}
H.~Caesar, V.~Bankiti, A.~H. Lang, S.~Vora, V.~E. Liong, Q.~Xu, A.~Krishnan,
  Y.~Pan, G.~Baldan, and O.~Beijbom.
\newblock nu{S}cenes: {A} multimodal dataset for autonomous driving.
\newblock {\em CoRR}, abs/1903.11027, 2019.

\bibitem{Chabot_2017_CVPR}
F.~Chabot, M.~Chaouch, J.~Rabarisoa, C.~Teuliere, and T.~Chateau.
\newblock Deep manta: A coarse-to-fine many-task network for joint 2d and 3d
  vehicle analysis from monocular image.
\newblock In {\em (CVPR)}, July 2017.

\bibitem{Chen_2016_CVPR}
X.~Chen, K.~Kundu, Z.~Zhang, H.~Ma, S.~Fidler, and R.~Urtasun.
\newblock Monocular 3d object detection for autonomous driving.
\newblock In {\em (CVPR)}, 2016.

\bibitem{NIPS2015_Chen}
X.~Chen, K.~Kundu, Y.~Zhu, A.~G. Berneshawi, H.~Ma, S.~Fidler, and R.~Urtasun.
\newblock 3d object proposals for accurate object class detection.
\newblock In {\em (NIPS)}, 2015.

\bibitem{Chen_2017_CVPR}
X.~Chen, H.~Ma, J.~Wan, B.~Li, and T.~Xia.
\newblock Multi-view 3d object detection network for autonomous driving.
\newblock In {\em (CVPR)}, July 2017.

\bibitem{Everingham2010}
M.~Everingham, L.~Van~Gool, C.~K.~I. Williams, J.~Winn, and A.~Zisserman.
\newblock The {P}ascal visual object classes {(VOC)} challenge.
\newblock {\em (IJCV)}, 88(2):303--338, 2010.

\bibitem{Geiger2012CVPR}
A.~Geiger, P.~Lenz, and R.~Urtasun.
\newblock Are we ready for autonomous driving? the kitti vision benchmark
  suite.
\newblock In {\em (CVPR)}, 2012.

\bibitem{He2017}
K.~He, G.~Gkioxari, P.~Doll{\'{a}}r, and R.~B. Girshick.
\newblock Mask {R-CNN}.
\newblock In {\em (ICCV)}, 2017.

\bibitem{He2015b}
K.~He, X.~Zhang, S.~Ren, and J.~Sun.
\newblock Deep residual learning for image recognition.
\newblock {\em CoRR}, abs/1512.03385, 2015.

\bibitem{TongHe_2019_arxiv}
T.~He and S.~Soatto.
\newblock Mono3d++: Monocular 3d vehicle detection with two-scale 3d hypotheses
  and task priors.
\newblock {\em CoRR}, abs/1901.03446, 2019.

\bibitem{Kehl_2017_ICCV}
W.~Kehl, F.~Manhardt, F.~Tombari, S.~Ilic, and N.~Navab.
\newblock Ssd-6d: Making rgb-based 3d detection and 6d pose estimation great
  again.
\newblock In {\em (ICCV)}, October 2017.

\bibitem{Kundu_2018_CVPR}
A.~Kundu, Y.~Li, and J.~M. Rehg.
\newblock {3D-RCNN}: Instance-level 3d object reconstruction via
  render-and-compare.
\newblock In {\em (CVPR)}, June 2018.

\bibitem{Lang_CVPR_2019}
A.~H. Lang, S.~Vora, H.~Caesar, L.~Zhou, J.~Yang, and O.~Beijbom.
\newblock Pointpillars: Fast encoders for object detection from point clouds.
\newblock In {\em (CVPR)}, 2019.

\bibitem{Law_2018_ECCV}
H.~Law and J.~Deng.
\newblock Cornernet: Detecting objects as paired keypoints.
\newblock In {\em (ECCV)}, September 2018.

\bibitem{cvpr19stereorcnn}
P.~Li, X.~Chen, and S.~Shen.
\newblock Stereo r-cnn based 3d object detection for autonomous driving.
\newblock In {\em (CVPR)}, 2019.

\bibitem{Liang_CVPR_2019}
M.~Liang, B.~Yang, Y.~Chen, R.~Hu, and R.~Urtasun.
\newblock Multi-task multi-sensor fusion for 3d object detection.
\newblock In {\em (CVPR)}, 2019.

\bibitem{Lin2016}
T.~Lin, P.~Doll{\'{a}}r, R.~B. Girshick, K.~He, B.~Hariharan, and S.~J.
  Belongie.
\newblock Feature pyramid networks for object detection.
\newblock {\em CoRR}, abs/1612.03144, 2016.

\bibitem{Lin+17}
T.~Lin, P.~Goyal, R.~B. Girshick, K.~He, and P.~Doll{\'{a}}r.
\newblock Focal loss for dense object detection.
\newblock {\em CoRR}, abs/1708.02002, 2017.

\bibitem{Liu+19}
L.~Liu, J.~Lu, C.~Xu, Q.~Tian, and J.~Zhou.
\newblock Deep fitting degree scoring network for monocular 3d object
  detection.
\newblock {\em CoRR}, abs/1904.12681, 2019.

\bibitem{Liu_2018_DetSurvey}
L.~Liu, W.~Ouyang, X.~Wang, P.~W. Fieguth, J.~Chen, X.~Liu, and
  M.~Pietik{\"{a}}inen.
\newblock Deep learning for generic object detection: {A} survey.
\newblock {\em CoRR}, abs/1809.02165, 2018.

\bibitem{Liu2016}
W.~Liu, D.~Anguelov, D.~Erhan, C.~Szegedy, S.~Reed, C.-Y. Fu, and A.~C. Berg.
\newblock Ssd: Single shot multibox detector.
\newblock In {\em (ECCV)}, 2016.

\bibitem{Manhardt_2019_CVPR}
F.~Manhardt, W.~Kehl, and A.~Gaidon.
\newblock Roi-10d: Monocular lifting of 2d detection to 6d pose and metric
  shape.
\newblock In {\em (CVPR)}, 2019.

\bibitem{Mousavian_2017_CVPR}
A.~Mousavian, D.~Anguelov, J.~Flynn, and J.~Kosecka.
\newblock 3d bounding box estimation using deep learning and geometry.
\newblock In {\em (CVPR)}, July 2017.

\bibitem{Zia_2014_CVPR}
K.~S. Muhammad Zeeshan~Zia, Michael~Stark.
\newblock Are cars just 3d boxes? jointly estimating the 3d shape of multiple
  objects.
\newblock In {\em (CVPR)}, 2014.

\bibitem{Murthy_17_ICRA}
J.~K. {Murthy}, G.~V.~S. {Krishna}, F.~{Chhaya}, and K.~M. {Krishna}.
\newblock Reconstructing vehicles from a single image: Shape priors for road
  scene understanding.
\newblock In {\em (ICRA)}, 2017.

\bibitem{Pillai_2019_ICRA}
S.~Pillai, R.~Ambrus, and A.~Gaidon.
\newblock Superdepth: Self-supervised, super-resolved monocular depth
  estimation.
\newblock In {\em (ICRA)}, 2019.

\bibitem{Qi_2018_CVPR}
C.~R. Qi, W.~Liu, C.~Wu, H.~Su, and L.~J. Guibas.
\newblock Frustum pointnets for 3d object detection from rgb-d data.
\newblock In {\em (CVPR)}, June 2018.

\bibitem{qin2019monogrnet}
Z.~Qin, J.~Wang, and Y.~Lu.
\newblock Monogrnet: A geometric reasoning network for 3d object localization.
\newblock In {\em (AAAI)}, 2019.

\bibitem{Redmon2016}
J.~Redmon, S.~Divvala, R.~Girshick, and A.~Farhadi.
\newblock You only look once: Unified, real-time object detection.
\newblock In {\em (CVPR)}, June 2016.

\bibitem{Redmon_2017_CVPR}
J.~Redmon and A.~Farhadi.
\newblock Yolo9000: Better, faster, stronger.
\newblock In {\em (CVPR)}, 2017.

\bibitem{Ren+15}
S.~Ren, K.~He, R.~Girshick, and J.~Sun.
\newblock Faster {R-CNN}: Towards real-time object detection with region
  proposal networks.
\newblock In {\em (NIPS)}, 2015.

\bibitem{Roddick18}
T.~Roddick, A.~Kendall, and R.~Cipolla.
\newblock Orthographic feature transform for monocular 3d object detection.
\newblock {\em CoRR}, abs/1811.08188, 2018.

\bibitem{RotPorKon18a}
S.~Rota~Bul\`o, L.~Porzi, and P.~Kontschieder.
\newblock In-place activated batchnorm for memory-optimized training of {DNN}s.
\newblock In {\em (CVPR)}, 2018.

\bibitem{Salton1986}
G.~Salton and M.~J. McGill.
\newblock {\em Introduction to Modern Information Retrieval}.
\newblock McGraw-Hill, Inc., New York, NY, USA, 1986.

\bibitem{shi2018pointrcnn}
S.~Shi, X.~Wang, and H.~Li.
\newblock Pointrcnn: 3d object proposal generation and detection from point
  cloud.
\newblock In {\em (CVPR)}, 2019.

\bibitem{Shin_arxiv_18}
K.~Shin, Y.~P. Kwon, and M.~Tomizuka.
\newblock Roarnet: {A} robust 3d object detection based on region approximation
  refinement.
\newblock {\em CoRR}, abs/1811.03818, 2018.

\bibitem{Wang_arxiv_2019}
Z.~Wang and K.~Jia.
\newblock Frustum convnet: Sliding frustums to aggregate local point-wise
  features for amodal 3d object detection.
\newblock {\em CoRR}, abs/1903.01864, 2019.

\bibitem{Wu_2017_CVPR_Workshops}
B.~Wu, F.~Iandola, P.~H. Jin, and K.~Keutzer.
\newblock Squeezedet: Unified, small, low power fully convolutional neural
  networks for real-time object detection for autonomous driving.
\newblock In {\em (CVPR) Workshops}, July 2017.

\bibitem{Xiang_2015_CVPR}
Y.~Xiang, W.~Choi, Y.~Lin, and S.~Savarese.
\newblock Data-driven 3d voxel patterns for object category recognition.
\newblock In {\em (CVPR)}, 2015.

\bibitem{Xiang_2017_WACV}
Y.~Xiang, W.~Choi, Y.~Lin, and S.~Savarese.
\newblock Subcategory-aware convolutional neural networks for object proposals
  and detection.
\newblock In {\em (WACV)}, 2017.

\bibitem{Xu_2018_CVPR}
B.~Xu and Z.~Chen.
\newblock Multi-level fusion based 3d object detection from monocular images.
\newblock In {\em (CVPR)}, June 2018.

\end{thebibliography}
}

\end{document}